%% file: main.tex
\documentclass[10pt]{article} 
\usepackage[accepted]{tmlr}

\input{math_commands.tex}

\usepackage{hyperref}
\usepackage{url}
\usepackage[capitalize,noabbrev]{cleveref}

\usepackage{microtype}
\usepackage{graphicx}
\usepackage{subcaption}
\usepackage{booktabs}

\usepackage{amsmath}
\usepackage{amssymb}
\usepackage{mathtools}
\usepackage{amsthm}
\usepackage{tabularx}
\usepackage{subcaption}
\usepackage{booktabs}

\usepackage[textsize=tiny]{todonotes}
\usepackage{pgfplots}

\usepackage{enumitem}
\usepackage{xspace}
\usepackage{wrapfig}
\usepackage{multirow}
\usepackage{pifont}

\usepackage{standalone}
\usepackage{tikz}
\usetikzlibrary{positioning}
\usetikzlibrary{arrows}
\usetikzlibrary{calc,fit}
\usetikzlibrary{shapes.geometric}
\input{imgs/diagrams/set2colors}

\graphicspath{
  {../}
}

\usepackage{algorithm}
\let\classAND\AND
\let\AND\relax
\usepackage{algorithmic}

\let\AND\classAND
\AtBeginEnvironment{algorithmic}{\let\AND\algoAND}

\newtheorem{lemma}{Lemma}

\crefname{section}{section}{sections}
\Crefname{section}{Section}{Sections}

\crefname{subsection}{section}{sections}
\Crefname{subsection}{Section}{Sections}

\title{On Designing Diffusion Autoencoders\\for Efficient Generation and Representation Learning}

\author{\name Magdalena Proszewska \email m.proszewska@ed.ac.uk \\
      \addr School of Informatics,
      University of Edinburgh
      \AND
      \name Esmeralda S. Whitammer \email esmeralda.whitammer@ed.ac.uk\\
      \addr School of Informatics,
      University of Edinburgh\\CIFAR Fellow, Learning in Machines and Brains
      \AND
      \name N. Siddharth \email n.siddharth@ed.ac.uk \\
      \addr School of Informatics,
      University of Edinburgh
}

\def\month{11}  
\def\year{2026} 
\def\openreview{\url{https://openreview.net/forum?id=KX0RAVVSWz}} 

\newcommand{\ours}{DMZ\xspace}

\begin{document}

\maketitle

\begin{abstract}
  Diffusion autoencoders (DAs) are variants of diffusion models that use an input-dependent latent variable to capture representations alongside the diffusion process.
  The latent variable is typically used for tasks such as downstream classification, controllable generation, and interpolation.
  However, the latent variable has an effect on unconditional generation, with performance strongly dependent on how this variable is modelled and learnt and how it is used in the denoising process.
  Here, we draw a connection between DAs and a particular class of improved diffusion models---those that learn their forward (noising) process---to argue that DAs have unrealised potential as generative models.
  We develop a variant of DAs, termed diffusion models with~\(z\) (\ours), to show that the adoption of certain design decisions, such as the choice of latent variables and conditioning method, can help realise this generative potential.
  Through extensive experiments and ablations, we show \ours{} converges in fewer iterations, generates high-quality samples with fewer denoising steps, and learns effective latent representations for downstream tasks like classification and domain transfer.
\end{abstract}

\section{Introduction}
\label{sec:intro}

Learning effective and efficient deep generative models with latent variables requires satisfying three aims: generating data that matches the observed data distribution, capturing relevant information in latent variables that facilitate interventions or downstream use, and doing both in a computationally efficient manner.

Diffusion models (DMs) are a powerful class of deep generative models that excel at generation, with deep hierarchical models \citep{ho2020denoisingdiffusionprobabilisticmodels} and score-based models \citep{song2021scorebasedgenerativemodelingstochastic} serving as foundations.
However, these models are computationally expensive and not designed to capture effective latent representations.
Approaches to address efficiency have largely focused on making generation faster and more robust \citep{song2022denoisingdiffusionimplicitmodels,nichol2021improveddenoisingdiffusionprobabilistic,ning2023inputperturbationreducesexposure,okhotin2023starshapeddenoisingdiffusionprobabilistic,shen2025efficient}.
Representation learning approaches using diffusion models have largely focused on extraction from pre-trained models, whether through latent codes \citep{zhang2023unsupervisedrepresentationlearningpretrained}, internal activations \citep{yang2023diffusionmodelrepresentationlearner,Xiang_2023_ICCV}, analysis of degradation patterns \citep{yue2024exploringdiffusiontimestepsunsupervised}, or by aiming to disentangle interpretable structures \citep{Tao2023DisDiff, zhang2023unsuperviseddiscoveryinterpretabledirections}.

Diffusion autoencoders (DAs) are a variant of DMs that incorporate input-dependent latent variables to capture representations alongside the diffusion process to enable reconstruction, controllable generation, and interpolations \citep{Preechakul_2022_CVPR,pmlr-v202-wang23ah}.
Their effectiveness at capturing such information, ability to subsequently generate data well, and potential to do so with fewer denoising steps all depend strongly on how well the latent variable is modelled and fit and how it is used in the denoising process.

Here, we draw a connection between DAs and approaches that learn the forward (noising) process, leading to more efficient learning and better generative models \citep{kingma2023variationaldiffusionmodels,bartosh2024neuralflowdiffusionmodels, nielsen2024diffencvariationaldiffusionlearned}, to argue that DAs can be better generative models.
We show that certain design decisions, including the choice and dimensionality of latent variable, method of conditioning the denoising process, and setup for learning the latent distribution allow us to obtain both efficient generation and meaningful latent representations.
We term this variant \ours{}; i.e., diffusion models with $z$, to reflect the fact that it only requires a latent variable \(z\) with no other components such as regularisers or constraints typical of DAs.

We show that \ours{} learns effective representations as evaluated on downstream tasks, including a novel domain-transfer setting for DAs, and is more efficient than standard DMs in two senses: it converges in fewer training iterations, and it reaches better sample quality with fewer denoising steps.

Our contributions are:
\begin{enumerate}[left=1ex,nosep]
\item  to show judiciously designed DAs converge faster and generate better samples in fewer denoising steps than standard DMs,
\item to explore and ablate design decisions that enable such efficiency, and
\item to evaluate representation quality through their latent variable, on a variety of tasks.
\end{enumerate}

\section{Background and related work}

Diffusion models (DMs) gradually corrupt data into noise through a forward process and learn to reverse this corruption. Denoising Diffusion Probabilistic Models \citep[DDPMs;][]{ho2020denoisingdiffusionprobabilisticmodels}
established this foundational setup with a Markovian noising process.

Given sample $x_0$ from the data distribution $q(x_0)$ and a predefined noise schedule $(\beta_1, \dots, \beta_T)$, the forward process simulates a Markov chain starting from data $x_0 \sim q(x_0)$, iteratively adding Gaussian noise over $T$ diffusion steps until obtaining a completely noisy image $x_T \sim \mathcal{N} (0, \mathbf{I})$ as
\vspace*{-0.5\baselineskip}
\begin{align*}
  q(x_{1:T} \mid x_0) &= \prod_{t=1}^{T} q(x_t \mid x_{t-1}), \qquad
  q(x_t \mid x_{t-1}) = \mathcal{N} \left( x_t; \sqrt{1 - \beta_t} x_{t-1}, \beta_t \mathbf{I} \right).
\end{align*}
Given observation \( x_0 \), the noised sample at \( t \) is derived as \(x_t = \sqrt{\bar{\alpha}_t} \, x_0 + \sqrt{1 - \bar{\alpha}_t} \, \epsilon\), where \(\epsilon \sim \mathcal{N}(0, \mathbf{I})\), \(\alpha_i = 1 - \beta_i\), and \(\bar{\alpha}_t =\prod_{i=1}^t \alpha_i\).
The reverse process employs a parametric denoiser $\mu_\theta$ as
\begin{align*}
  p_\theta(x_{t-1} | x_t)
  &= \mathcal{N} \left( x_{t-1}; \mu_\theta(x_t, t), \sigma_t^2 \mathbf{I} \right),
  \quad \text{where}\,\sigma_t = \sqrt{\frac{1 - \bar{\alpha}_{t-1}}{1 - \bar{\alpha}_t} \beta_t}.
\end{align*}
Instead of directly predicting the mean of the forward process posterior $ \mu_\theta(x_t, t)$, \citet{ho2020denoisingdiffusionprobabilisticmodels} propose training a neural network \( \epsilon_\theta(\cdot) \) to predict noise \( \epsilon \) by optimising
\begin{equation}
  \mathbb{E}_{x_0 \sim q(x_0),\, \epsilon \sim \mathcal{N}(0, \mathbf{I}),\, t \sim \mathcal{U}(\{1, \dots, T\})}
  \!\left[ \| \epsilon - \epsilon_\theta(x_t, t) \|^2 \right].
  \label{eq:loss}
\end{equation}
The noise predictor $\epsilon_\theta$ minimising \eqref{eq:loss} can be expressed in terms of the score of the marginal distribution of $x_t$, establishing a connection between denoising autoencoders and score-based modelling \citep{song2021scorebasedgenerativemodelingstochastic}.
The reverse (denoising) process is then defined in terms of \(\mu_\theta\) taken as
\begin{align*}
  \mu_\theta(x_t, t)
  &= \frac{1}{\sqrt{\alpha_t}} \left( x_t - \frac{1 - \alpha_t}{\sqrt{1 - \bar{\alpha}_t}} \epsilon_\theta(x_t, t) \right)
\end{align*}
Markovian DMs were later extended to non-Markovian variants \citep{song2022denoisingdiffusionimplicitmodels,okhotin2023starshapeddenoisingdiffusionprobabilistic}, where input $x_0$ influences the denoising process, requiring fewer steps for inference.
These models assume a fixed forward process and focus solely on learning the reverse denoising process.

\paragraph{Diffusion models with learned forward process:}

Recent work explores parametrising and learning the forward process (noising) as well as the denoising process.
VDMs \citep{kingma2023variationaldiffusionmodels}, NFDMs \citep{bartosh2024neuralflowdiffusionmodels} and DiffEnc \citep{nielsen2024diffencvariationaldiffusionlearned} learn both the forward process $q_\phi(x_t|x_0, t)$ and the reverse process $p_\theta(x_0 | x_t)$, and have been shown to achieve better log-likelihood, potentially requiring fewer steps for inference.
Other work explores conditional diffusion and use of data-dependent priors \citep{lee2022priorgradimprovingconditionaldenoising} or shifts \citep{zhang2023shiftddpmsexploringconditionaldiffusion}.
This direction parallels the motivations behind hierarchical variational autoencoders (VAEs) \citep{vahdat2021nvaedeephierarchicalvariational, kuzina2024hierarchicalvaediffusionbasedvampprior}, which introduce multi-level latent structures to better capture data distributions.

\looseness=-1
\paragraph{Diffusion autoencoders (DAs):}
This class of models introduces a latent variable that guides denoising, enabling tasks such as retrieval and editing through learned representations \citep{Preechakul_2022_CVPR,pmlr-v202-wang23ah,pandey2022diffusevaeefficientcontrollablehighfidelity,hudson2023sodabottleneckdiffusionmodels}.
Note that DAs are also effectively non-Markovian DMs as they condition denoising on the target $x_0$ through the latent variable.
All instances of DAs excepting \citet{hudson2023sodabottleneckdiffusionmodels} tackle unconditional generation, which aligns with the focus of our work.
DiffAE \citep{Preechakul_2022_CVPR} employs an encoder $z=\text{Enc}_\phi(x_0)$ whose output is used at each step of denoising alongside $x_t$ and $t$.
InfoDiffusion \citep{pmlr-v202-wang23ah}, based on InfoVAE \citep{zhao2018infovaeinformationmaximizingvariational}, further introduces a probabilistic encoder to maximise MI and align the posterior with a discrete prior of $z$.
DiffuseVAE \citep{pandey2022diffusevaeefficientcontrollablehighfidelity} combines VAE and DDPM in a two-stage process, with a DDPM conditioned on the reconstructions of a pre-trained VAE.
While these models demonstrate the ability to control the denoising process via a learned latent variable, they share a key limitation in terms of their generative performance.
At inference time, they all rely on auxiliary samplers—such as DDIMs \citep{Preechakul_2022_CVPR,pmlr-v202-wang23ah} or GMMs \citep{pandey2022diffusevaeefficientcontrollablehighfidelity}—to produce valid latent codes, introducing unnecessary overhead.

\section{Diffusion models with \(z\): \ours{}}

\begin{wrapfigure}[27]{r}{0.46\textwidth}
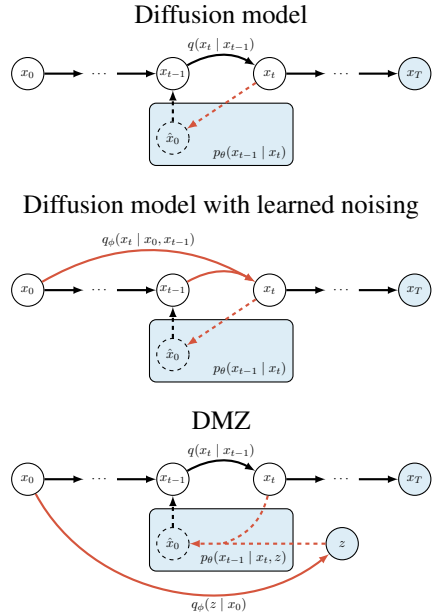

\vspace{-3.75\baselineskip}
  \centering
  Diffusion model\\[0.25\baselineskip]
  \includestandalone[width=0.95\linewidth]{imgs/diagrams/diff-hvi}\\[0.25\baselineskip]
  Diffusion model with learned noising\\[0.25\baselineskip]
  \includestandalone[width=0.95\linewidth]{imgs/diagrams/diff-learnable}\\[0.25\baselineskip]
  \ours\\[0.25\baselineskip]
  \includestandalone[trim=24 12 0 0,width=0.95\linewidth]{imgs/diagrams/diff-dmz}\\[0.25\baselineskip]
  \caption{\textbf{Top:} Diffusion reverses a Markovian noising process from \(x_T\) (possibly via predicted $x_0$ at each step).
    \textbf{Middle:} A generalisation where generation reverses a non-Markovian learned noising process, marginalising out unknown $x_0$.
    \textbf{Bottom:} \ours, where generation conditions on $z$.
    %
    Solid and dashed arrows denote noising and generation respectively.
    Red arrows denote learned parametric models.
    Blue objects denote data necessary for generation: initial noise, transition kernel, and $z$.
  }
  \label{fig:models}
\end{wrapfigure}
A DA with a stochastic encoder $q_\phi(z\mid x_0)$ is trained with a loss that generalises \eqref{eq:loss}:
\begin{equation}
  \mathbb{E}_{\substack{x_0 \sim q(x_0),\, \epsilon \sim \mathcal{N}(0, \mathbf{I}),\\ t \sim \mathcal{U}(\{1, \dots, T\}),\,z\sim q_\phi(z\mid x_0)}}
  \left[ \| \epsilon - \epsilon_\theta(x_t, t,z) \|^2 \right].
\end{equation}

Unlike prior approaches that employ additional terms to constrain the latent variables, \ours{} directly optimises this objective with respect to the denoiser $\epsilon_\theta$ (now conditioned on $z$) and the encoder $q_\phi$.

Here, we describe a specific variant of diffusion autoencoders (DAs), following a set of judicious design choices that allow both for efficient generative modelling and capturing effective latent representations.
These choices are not independent. Each follows from the goal to capture useful information in $z$, turning an unconditional denoiser into a conditional one, reducing variance and improving denoised samples (\cref{app:proof}).
A latent alone is insufficient. To our knowledge, prior DAs have not shown generative gains from latent representations, and our ablations highlight the importance of latent size and conditioning.
We first draw attention to the key distinctions between DMs, DMs with a learned forward (noising) process, and a particular type of DA, in \cref{fig:models}.

Diffusion models with a learnable forward process (middle) construct a noised observation \(x_t\) by incorporating side-information from the observation \(x_0\), through a learnable parametric function.
This means the source for the denoiser, \(x_t\), can carry additional information to help denoise better to~\(x_{t-1}\).
This is also a fundamental feature of DAs (bottom)---they incorporate side-information through~\(z\) into the denoising process by additionally conditioning the denoiser on \(z\).
In effect, denoising in DAs can be seen as \(\{x_t, z\} \mapsto x_{t-1}\), with \(x_t\) derived through a standard fixed noising process, just that the information from this \(x_t\) and \(z\) are not explicitly combined and required to additionally satisfy the constraints of the noising process.
This also means that one needs to also be able to sample from the latent \(z\) in order to function as a proper generative model; this is what the rest of our design choices seek to address.

\paragraph{Choice of latent \(z\):}
In choosing the type of latent variables, we note the importance of discrete latent variables for representation learning.
They offer a more interpretable and more space-efficient way to represent data compared to continuous latent variables, and can also capture structured relationships and details, leading to simpler and more effective models \citep{rolfe2017discretevariationalautoencoders,vahdat2018dvaediscretevariationalautoencoders,oord2018neuraldiscreterepresentationlearning,razavi2019generatingdiversehighfidelityimages,kaiser2018discreteautoencoderssequencemodels,metz2019discretesequentialpredictioncontinuous}.
%
Their type (discrete or continuous) and dimensionality form the basis of the experiments (\Cref{sec:experiments}) and ablations (\Cref{sec:ablations}) we perform.

With discrete $z$, we explore the choice of \emph{binary} latent variables following recent evidence showcasing their effectiveness in the diffusion and reinforcement learning settings \citep{meyer2024harnessing,wang2023binary}.
%
With regard to dimensionality of the latent variable, generally speaking, a lower-dimensional latent is likely to be more interpretable.
%
Conversely, a larger number of dimensions is capable of capturing more information, helping with use in downstream tasks.
An additional consideration comes from how the size of the latent variable affects our ability to model a prior over the latent~\(z\) to allow sampling at inference time.

\begin{figure}[!t]
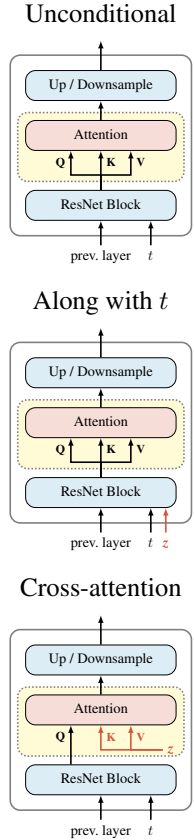

  \centering
  \begin{tabular}{c c c}
    Unconditional & Along with $t$ & Cross-attention \\ 
    \includestandalone[width=0.2\linewidth]{imgs/diagrams/uncond}
    &
    \includestandalone[width=0.2\linewidth]{imgs/diagrams/condwitht}
    &
    \includestandalone[width=0.2\linewidth]{imgs/diagrams/condatt}
  \end{tabular}
  \caption{\textbf{Left:} Denoiser with time conditioning and optional attention. \textbf{Mid, right:} Two conditioning strategies with $z$.}
  \label{fig:cond}
\end{figure}

\paragraph{Conditioning on latent \(z\):}
Another key design decision for effective modelling is the manner in which the denoiser is conditioned on $z$.
In the standard (DM) case, the denoiser simply takes the noisy observation~\(x_t\) along with an encoding of time step~\(t\).
Since existing diffusion autoencoder approaches have been developed and evaluated primarily using U-Net backbones, we adopt the same architecture.
A typical model architecture for such conditioning is shown in \cref{fig:cond} (left), with the time step given as input to each block of a denoising U-Net, where some blocks include self-attention.
A natural way to extend this to condition on~\(z\) would be to include it along with \(x_t\) and~\(t\) as shown in \cref{fig:cond} (middle), as done in some prior DAs \citep{Preechakul_2022_CVPR,pmlr-v202-wang23ah}.
%
However, such global conditioning can entangle the influence of $z$ across all computation paths, potentially encouraging shortcut solutions and reducing selective use of $z$.

As an alternative, we suggest that \(z\) can be more effective when used \emph{only} in modulating attention, and so use \emph{cross-attention} for some blocks with attention layers---with keys \(K\) and values \(V\) coming from \(z\), and queries \(Q\) coming from the original inputs to the denoiser.
The rationale is that cross attention allows a parametric function (denoiser) to more flexibly reflect conditioning by transforming the function parameters.
Concatenation would only affect conditioning indirectly via the outputs of different layers.
As we will see in the experiments, this is a useful inductive bias and can have a marked effect on learning effective and useful representations.
While we primarily use a U-Net backbone, the \ours{} framework and its cross-attention conditioning strategy extend naturally to Diffusion Transformers \citep{Peebles2022DiT} (see \Cref{app:dit}).

\paragraph{Learning with latent \(z\):}
An interesting feature of DAs is that the latent variable \(z\) is not required to attain the optimum.
There is no specific pressure for the model to capture \emph{any} information in \(z\) given that the standard noising and denoising processes are sufficiently flexible to faithfully model and generate observed data.
This lies in direct contrast to typical deep generative models that employ their latent variables as a \emph{bottleneck}, forcing all information through it.

However, using \(z\) effectively helps better optimise the training objective as it seeks to reconstruct the target observation (or noise).
We derive this condition in \Cref{app:proof}, showing that
\[
  \mathcal{L}_{\text{DMZ}} = \mathbb{E}[\operatorname{Var}(\epsilon \mid x_t,z)] \le \mathbb{E}[\operatorname{Var}(\epsilon \mid x_t)] = \mathcal{L}_{\text{DDPM}}.
\]

This concerns the optimal estimators only. It shows that $z$ cannot hurt the ideal objective. It does not imply faster optimisation or fewer sampling steps. Those are our empirical findings.

The inequality holds for the unconstrained estimator of $z$. When $z$ is learned via a variational objective, the KL regularisation term penalises information encoded in the latent, making it relatively easy to satisfy constraints that encourage $q(z\mid x)$ to match a non-informative prior like $\mathcal{N}({0}, \boldmath{I})$ (see \Cref{app:kldiv}). As a result, the model may ignore $z$ if the predictive gain does not offset the KL cost, leading to posterior collapse and a non-informative latent representation. This phenomenon has been observed in prior work \citep{pmlr-v202-wang23ah}, where additional mutual information regularisation was used to encourage latent usage. Other approaches avoid this issue by employing pretrained probabilistic models with well-defined priors \citep{pandey2022diffusevaeefficientcontrollablehighfidelity}. Standard variational regularisation is thus not well-suited here because $z$ is not required.

That $z$ is not required also means that usual intuitions about the prior's flexibility and capacity do not hold---it only needs to capture information used by the denoiser. As we show experimentally, binary latents suffice. 
For small $|z|$, the support size $2^{|z|}$ is comparable to the number of samples drawn from $q_\phi(z\mid x_0)$ across the dataset and the training timesteps (e.g., $2^{16}=65{,}536$ for $|z|=16$).
In this regime, the denoiser encounters nearly all possible values during training, so sampling $z\sim\mathrm{Bernoulli}(0.5)$ at inference time encounters little to no unseen values, while $z$ can still encode meaningful information.
Keeping $|z|$ small is thus a deliberate design choice that enables sampler-free inference: we simply sample $z$ from a uniform prior at inference time.

Put together, we find that judicious design choices allow us to construct a model that (a) does not need auxiliary losses, (b) does not need additional learning of the prior, (c) captures effective representations, and (d) can do all this while being faster to learn than standard diffusion models.

\section{Experiments}
\label{sec:experiments}

Our experiments are organised around a single question: does conditioning the denoiser on a latent $z$ make diffusion models more efficient to train and sample from, while keeping $z$ simple to sample from?
Primary evidence is in \cref{sec:impactz} (training convergence and use of $z$ during denoising) and \cref{sec:gen} (generation quality at few denoising steps, with $z$ drawn from a simple prior).
All other evaluations support or extend this: \cref{sec:rep} shows that $z$ captures meaningful representations, \cref{sec:ablations} ablates how each design choice serves the goal, and \cref{sec:multi} demonstrates utility of $z$ for other tasks.

All our models are trained following the setup of \citet{nichol2021improveddenoisingdiffusionprobabilistic}, using their architecture and training hyperparameters.
More details can be found in the Appendix and the code\footnote{\url{https://github.com/exlab-research/dmz}}.
Throughout, $T$ in the experiments denotes the number of denoising steps used at inference, which is independent of the training horizon: all models are trained with $T_{\text{train}}=1000$, and at inference we mainly use $T_{\text{infer}}\in\{10,20,50,100\}$ timesteps spaced uniformly over $\{1,\dots,1000\}$.
We train until no further improvement in FID scores \citep{Seitzer2020FID} is observed for $T=100$ denoising steps.
We denote \ours{}-$n$ as an instance of \ours{} with a latent dimensionality of $|z| = n$.
The dimensionality of $z$ was deliberately kept small, guided by the number of available labels in each dataset and the requirements of downstream tasks; this choice is further examined in the ablation study.
Additional results and samples are provided in the Appendix.

\subsection{Impact of latent $z$}
\label{sec:impactz}

\begin{figure}[t!]
    \centering
    \begin{subfigure}[b]{0.49\linewidth}
        \centering
        \includegraphics[trim=5mm 6mm 5mm 5mm, clip, width=0.49\linewidth]{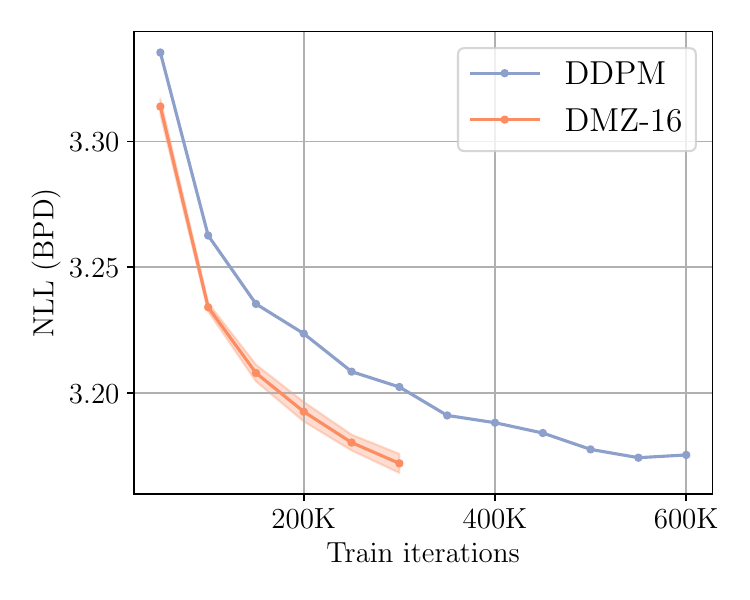}
        \includegraphics[trim=5mm 6mm 5mm 5mm, clip, width=0.49\linewidth]{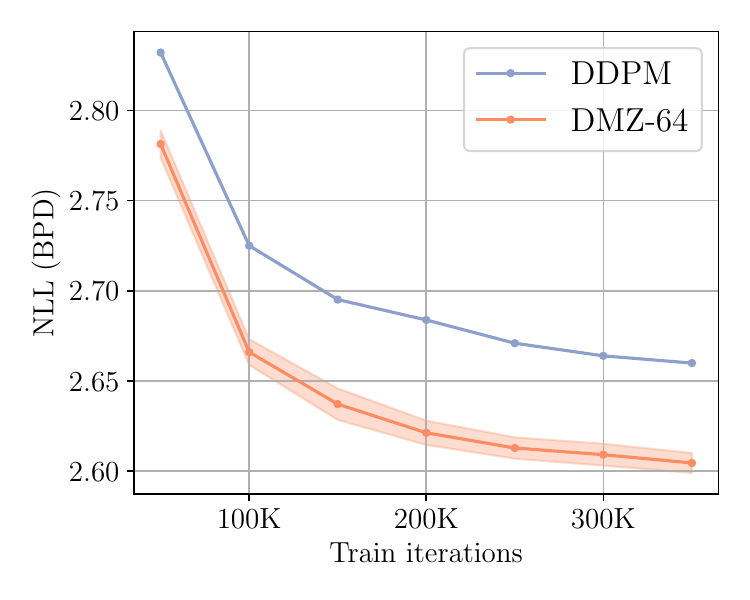}
        \caption{NLL for CIFAR-10 (l) and CelebA-64 (r)}
    \end{subfigure}
    \begin{subfigure}[b]{0.49\linewidth}
        \centering
        \includegraphics[trim=5mm 6mm 5mm 5mm, clip, width=0.49\linewidth]{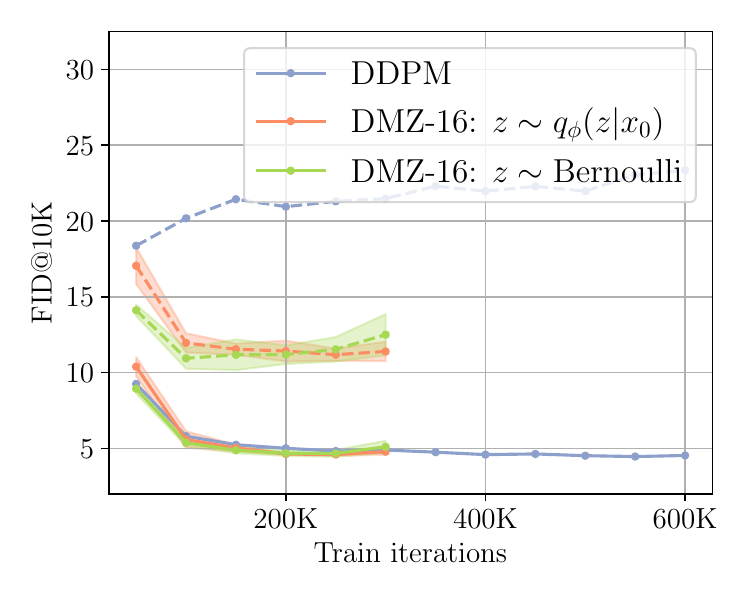}
        \includegraphics[trim=5mm 6mm 5mm 5mm, clip, width=0.49\linewidth]{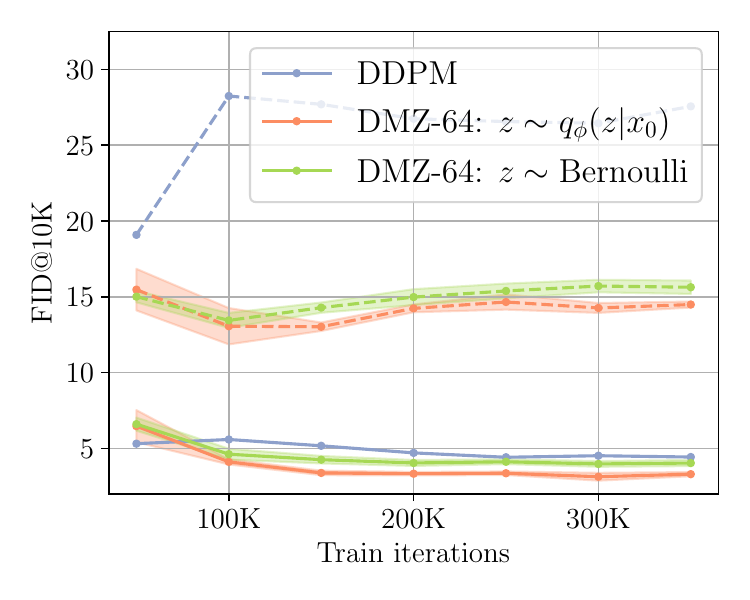}
        \caption{FID for CIFAR-10 (l) and CelebA-64 (r)}
    \end{subfigure}
    \caption{Comparison of training curves for \ours and the baseline DDPM. Dashed lines correspond to results for $T=10$, while solid lines indicate $T=100$. For \ours{}, we plot mean$\pm$std over 5 runs.}
    \label{fig:train}
    \vspace{0.5\baselineskip}
\end{figure}

\paragraph{On training and efficiency:}
We evaluate the impact of the latent variable $z$ on training efficiency by comparing \ours{} to its unconditional counterpart---a standard DDPM.
This baseline shares the same architecture and training procedure as \ours{}, differing solely in the absence of $z$-specific components.
Following prior work, we train models on CIFAR-10 \citep{Krizhevsky09learningmultiple} and CelebA-64 \citep{liu2015faceattributes}.
We report FID scores \citep{heusel2018ganstrainedtimescaleupdate,Seitzer2020FID} and negative log-likelihood (NLL) in bits per dimension.

As shown in \cref{fig:train}, \ours{} converges faster and achieves better generation efficiency. Notably, for $T=10$, it achieves much lower FID scores, demonstrating improved performance with fewer denoising steps.
The gain at small $T$ is not only faster learning; \ours{} also reaches a better asymptotic FID than DDPM.
We attribute this to \ours{} targeting a better objective than the unconditional DDPM. When $z$ captures useful information about $x_0$, the conditional denoiser is more flexible, so it can better fit the \ours{} objective and each denoising step becomes easier to model, requiring fewer steps.
This is the same mechanism that makes conditional diffusion effective, and it parallels DDIM \citep{song2022denoisingdiffusionimplicitmodels}, which also conditions on information about $x_0$ for non-Markovian generation. In DDIM this is a prediction of $x_0$ from $x_t$, whereas in \ours{} it is a stochastic encoding $z \sim q_\phi(z \mid x_0)$.
We also observe that a lower NLL does not necessarily correspond to better FID scores, highlighting the often-misaligned objectives of likelihood maximisation and perceptual sample quality.
Our work aligns with prior efforts to improve sampling quality and efficiency in DDPMs, rather than optimising NLL.

\begin{wrapfigure}[12]{r}{0.3\linewidth}
\vspace{-2\baselineskip}
  \centering
  \includegraphics[trim=5mm 6mm 5mm 5mm, clip, width=0.95\linewidth]{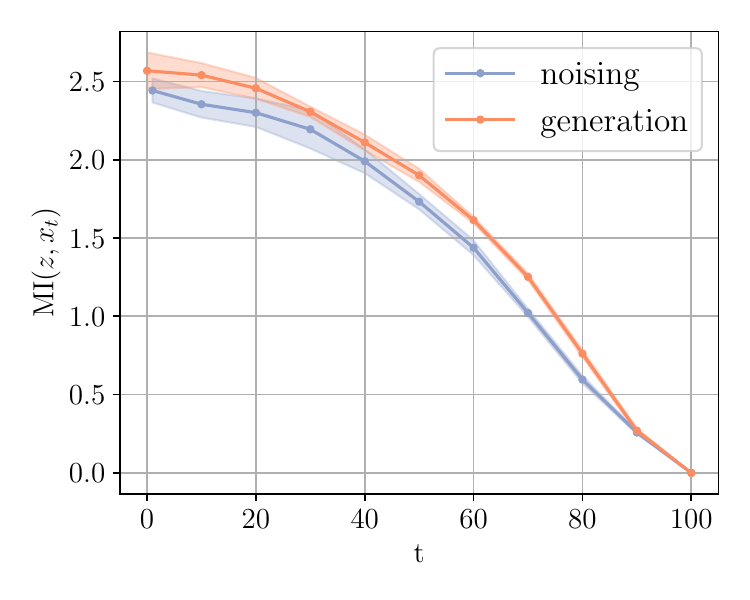}
  \caption{MI between $z$ learned by DMZ and $x_t$ from the noising process (blue) and from the denoising process (orange; starting from $\{x_T, z\}, x_T \sim \mathcal{N}(0, \mathbf{I})$).}
  \label{fig:mi}
\end{wrapfigure}

\paragraph{On the denoising process:}
We now examine the role of $z$ at different stages of the denoising process (generation) by measuring the mutual information (MI) between the learned representations $z \sim q_\phi(z \mid x_0)$ and (a) noised data $x_t \sim q(x_t \mid x_0, t)$, and (b) denoised generations $x_t \sim p_\theta(x_t \mid x_{t+1},z),\; x_T\sim \mathcal{N}(0,\mathbf{I})$.

For DMZ-16 (\(\lvert z\rvert=16\)) trained on CIFAR-10, we fix $T=100$, and compute MI between sampled $z$ and $x_t$ ($t$ taking 11 evenly spaced values) 
using MINE~\citep{belghazi2021minemutualinformationneural}\footnote{\url{https://github.com/gtegner/mine-pytorch}}. \cref{fig:mi} illustrates how MI evolves over time.
We find that while $z$ is not required given $x_t$ during training, there is non-negligible MI between them, indicating that the network learns to extract meaningful information from $z$. Furthermore, $z$ provides information consistently throughout the denoising process.

\subsection{Generation quality and efficiency}\label{sec:gen}

We show that \ours{} outperforms existing DAs at most settings, while remaining competitive with DMs, particularly at small step budgets.
Following prior work, we evaluate \ours{}-16 and \ours{}-64 on CIFAR-10 and CelebA-64 respectively, using FID scores (FID@50K) across various inference step counts ($T=10,20,50,100$).
Results are presented in \cref{tab:fid}, and include scores obtained by an unconditional counterpart to \ours{}, DDPM, which serves as a natural baseline as discussed in \cref{sec:impactz}.
For fair comparison, we also report scores with 10K samples in \Cref{app:fid10}. Inception scores \citep{3157096.3157346} are in \Cref{tab:is}.

\ours{} achieves excellent FID scores on both datasets. These scores, even with fewer inference steps, highlight efficiency at generation.
Here, efficiency refers to better FID with fewer denoising steps and faster training convergence, not lower cost per step. Nominally the cross-attention blocks add per-step overhead. Comparisons with capacity-matched baselines, sampling times, and training costs are in \Cref{app:efficiency}.
Comparing \ours{} with the DDPM, we again show (\Cref{sec:impactz}) the benefit provided by including $z$.

\begin{wraptable}[30]{r}{0.5\linewidth}
\vspace{-1.5\baselineskip}
  \centering
  \caption{FID scores. All DAs except DiffAE use DDPMs (DiffAE only available as DDIM).
   Models marked $^*$ used 10K samples; all others used 50K. \ours{} reports mean and stdev over 5 seeds.}\label{tab:fid}
   \footnotesize
    \begin{tabular}{@{}l@{}rrr}
        \toprule
         Model & $T$ & CIFAR-10 & CelebA-64 \\
        \midrule
        \multirow{4}{*}{\ours} & 10 & 9.78$\pm$0.35  &  14.43$\pm$0.45 \\
        & 20 & 4.79$\pm$0.40 & 7.89$\pm$0.19 \\
        & 50 & 3.04$\pm$0.28 & 4.20$\pm$0.16 \\
        & 100 & 2.79$\pm$0.19 & 2.93$\pm$0.10 \\
        \midrule
        \multirow{4}{*}{\shortstack[l]{DDPM\\ \citep{ning2023inputperturbationreducesexposure} \\(reproduced)}} & 10 & 21.25 & 25.72 \\ 
        & 20 & 7.21 & 13.13 \\
        & 50 & 3.18 & 5.69 \\
        & 100 & 2.56 & 3.34 \\
        \midrule
        \multirow{4}{*}{\shortstack[l]{DiffAE \\ \citep{Preechakul_2022_CVPR}}} & 10  & --- & 12.92 \\
        & 20 & --- & 10.18  \\
        & 50 & --- & 7.05 \\
        & 100 & --- & 5.30 \\
        \midrule
        \multirow{2}{*}{\shortstack[l]{InfoDiffusion$^*$ \\\citep{pmlr-v202-wang23ah}}} & \multirow{2}{*}{1000} & \multirow{2}{*}{31.5} & \multirow{2}{*}{21.2} \\ \\
        \midrule
        \multirow{4}{*}{\shortstack[l]{DiffuseVAE$^*$ \\\citep{pandey2022diffusevaeefficientcontrollablehighfidelity}}} & 10  & 34.22 & 25.79 \\
        & 25 & 17.36 & 13.89 \\
        & 50 & 11.00 & 9.09 \\
        & 100 & 8.28 & 7.15 \\
        \midrule
        \multirow{2}{*}{\shortstack[l]{VDM \\ \citep{kingma2023variationaldiffusionmodels}}} & \multirow{2}{*}{1000} & \multirow{2}{*}{4.0} & \multirow{2}{*}{---} \\ \\
        \midrule
        \multirow{2}{*}{\shortstack[l]{DiffEnc$^*$ \\\citep{nielsen2024diffencvariationaldiffusionlearned}}} & \multirow{2}{*}{1000} & \multirow{2}{*}{14.6} & \multirow{2}{*}{---} \\ \\
        \midrule
        \multirow{3}{*}{\shortstack[l]{NFDM \\ \citep{bartosh2024neuralflowdiffusionmodels}}} & 2 & 12.44 & --- \\
        & 4 & 7.76 & --- \\
        & 12 & 5.2 & ---\\
        \midrule
        \multirow{5}{*}{\shortstack[l]{DDIM \\\citep{song2022denoisingdiffusionimplicitmodels}}} & 10 & 13.36 & 17.33 \\
        & 20 & 6.84 & 13.73 \\
        & 50 & 4.67 & 9.17 \\
        & 100 & 4.16 & 6.53 \\
        & 1000 & 4.04 & 3.51 \\
        \bottomrule
    \end{tabular}
\end{wraptable}

Note that unlike baseline DAs which rely on auxiliary samplers for $z$, \ours\ samples $z$ directly from a Bernoulli prior with no auxiliary sampler and no extra sampling cost for $z$.
This is possible as targeting both informative posteriors and a uniform prior is not contradictory: ideally, each individual posterior captures targeted information, while the \emph{aggregate} posterior is spread out or uniform.
We find evidence of this in a \ours{}-16 model trained on CIFAR-10.
The expected posterior entropy, $\mathbb{E}_{q(x_0)}\!\left[H[q_\phi(z\mid x_0)]\right] = 0.1191$ nats, is very low, indicating that the posteriors are sharply peaked and capture useful information about $x_0$.
At the same time, $\mathrm{KL}\!\left(q_\phi(z)\,\|\,p(z)\right) = 0.0006$ nats, 
indicating that the per-dimension marginals of the aggregate posterior are very close to uniform Bernoulli prior. See \Cref{app:proof} for a mispaired-$z$ control experiment.
KL is estimated from per-dimension marginals using 10K samples.
DiffAE achieves better FID on the CelebA-64 dataset for $T=10$, which we attribute to its use of DDIM as the base generative model.
We show results combining \ours{} with DDIM in \Cref{app:ddim}.
Overall, \ours{} demonstrates stronger performance and is simpler compared to prior DAs.

\ours{} achieves comparable performance to state-of-the-art DMs, particularly when accounting for the number of denoising steps.
Moreover, we impose minimal architectural constraints, unlike NFDM, which has constraints that limit scalability and flexibility.

\subsection{Representations}\label{sec:rep}

\begin{wraptable}[12]{r}{0.5\linewidth}
\vspace{-0.5\baselineskip}
    \centering
    \caption{Assessment of representation quality via downstream classification tasks. Mean and stdev computed across 5-fold splits.}
  \label{tab:acc}
  \small
    \begin{tabular}{@{}l@{\quad}cccc}
      \toprule
      Dataset $\rightarrow$ & \multicolumn{2}{c}{CIFAR-10} & \multicolumn{2}{c}{CelebA-64}  \\
      \cmidrule(lr){2-3}\cmidrule(lr){4-5}
      Model $\downarrow$ & $\lvert z \rvert$ & Acc & $\lvert z \rvert$ & AUROC \\
      \midrule
      DiffAE & 32 & 39.5 & 32 & 79.9 \\
      InfoDiffusion & 32 & 41.2 & 32 & 84.8 \\
      \midrule
      \multirow{4}{*}{\ours} & 16 & 39.50$\pm$0.47 & 32 & 84.85$\pm$0.03  \\
                            & 32 & 42.35$\pm$0.56 &  64 & 86.43$\pm$0.05 \\
                            & 64 & 46.45$\pm$0.46 &  128 & 87.89$\pm$0.02  \\
                            & 128 & 49.71$\pm$0.50 & 256 & 88.48$\pm$0.04 \\
      \bottomrule
    \end{tabular}
\end{wraptable}

\paragraph{Quantitative evaluation:}

We assess the quality of the learned representations $z$ following \citet{pmlr-v202-wang23ah}, by measuring classification accuracy of a logistic classifier trained on the latents. For each dataset, we extract $z$ for all data and report classification performance using 5-fold cross-validation.
For CIFAR-10, we report classification accuracy. For CelebA-64, we report the average AUROC over 40 binary attributes to account for class imbalance.

\Cref{tab:acc} relates $\lvert{}z\rvert$ to classification accuracy. Our results, alongside those of \citet{pmlr-v202-wang23ah}, show that \ours{} performs on par or better than InfoDiffusion without explicit constraints or regularisation.

\ours{}'s core idea is to learn representations that guide the denoiser to generate more efficiently. In assessing their effectiveness, while we do not expect state-of-the-art accuracy on downstream tasks, results (\Cref{tab:acc}) show the representations are meaningful and compare favourably to our DA baselines. \Cref{sec:multi} further explores their utility on a multimodal setup using these representations.

\paragraph{Qualitative evaluation:}

\begin{wrapfigure}[13]{r}{0.45\linewidth}
\vspace{-\baselineskip}
  \centering
  \includegraphics[trim=134 3 0 1, clip, width=\linewidth]{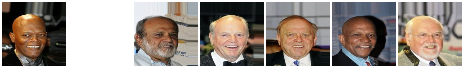}
  \includegraphics[trim=134 3 0 1, clip, width=\linewidth]{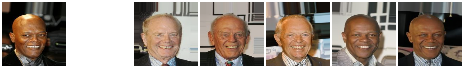}
  \includegraphics[trim=134 3 0 1, clip, width=\linewidth]{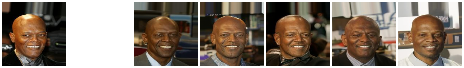}
  \caption{\ours{} exemplar generations sharing the same \(z\). Rows correspond to $|z|=64, 128, 256$.}
  \label{fig:encode}
\end{wrapfigure}
%
For each model, we sample data, $x_0 \sim \mathcal{D}$, and its corresponding latent~\(z\). For this \(z\), we generate by sampling different noise vectors $x_T \sim \mathcal{N}(0, \mathbf{I})$ and denoise via $p_\theta(x_T, z)$.
\cref{fig:encode} presents examples illustrating the impact of the size of $z$ on representations.
We observe that for smaller $|z|$, less information about the image is retained. Low-level attributes, such as the presence of a smile, are preserved, while higher-level features, such as ethnicity, are not consistently captured. We attribute this to the fact that high-level features remain present in the intermediate representations $x_t$ passed to the network alongside $z$. As $|z|$ increases, the model is able to encode more information, including higher-level semantic attributes, resulting in generated images that remain consistent across different samples of $x_T$.

\begin{figure}[t]
  \centering
  \begin{subfigure}{0.49\linewidth}
    \centering
    \includegraphics[width=\linewidth]{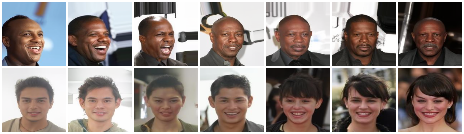}
    \caption{Latent interpolation using \ours{}-256.}
    \label{fig:interp}
  \end{subfigure}
  \begin{subfigure}{0.49\linewidth}
    \centering
    \begin{tikzpicture}
      \node[anchor=south west,inner sep=0] (image) at (0,0) {
        \includegraphics[width=\linewidth]{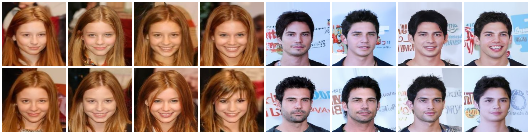}
      };
      \begin{scope}[x={(image.south east)},y={(image.north west)}]
        \node[font=\scriptsize] at (0.06, 1.1) {smile};
        \node[font=\scriptsize] at (0.19, 1.1) {mouth};
        \node[font=\scriptsize, yshift=-0.4mm] at (0.32, 1.1) {age};
        \node[font=\scriptsize, yshift=-0.2mm] at (0.44, 1.1) {bangs};
        \node[font=\scriptsize] at (0.56, 1.1) {beard};
        \node[font=\scriptsize] at (0.69, 1.1) {smile};
        \node[font=\scriptsize, yshift=-0.4mm] at (0.82, 1.1) {age};
        \node[font=\scriptsize] at (0.95, 1.1) {mouth};
    \end{scope}
    \end{tikzpicture}
    \caption{Targeted edits using \ours{}-256 and CelebA attributes. 
}
    \label{fig:edits}
  \end{subfigure}
  \caption{Qualitative evaluation on CelebA-64.}
  \label{fig:celeba}
\end{figure}
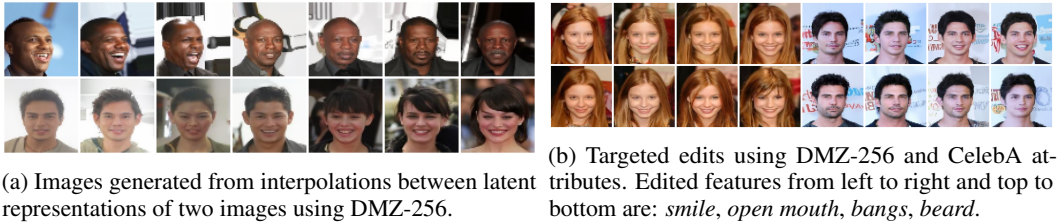

Next, we illustrate the properties of the learned representations $z$ through interpolation. 
Specifically, for a given pair of images we extract their corresponding latents, $z_{\text{source}}$ and $z_{\text{target}}$ and interpolate in discrete steps by sequentially flipping bits in $z_{\text{source}}$ to match those in $z_{\text{target}}$. See \cref{alg:inter} in \cref{app:inter-edit} for further details.
\cref{fig:interp} presents examples for \ours{}-256 trained on CelebA-64.

Moreover, via $z$, we can perform targeted edits on the generated samples---altering attributes like hair or facial expression---using classifiers trained on the latent representations. We use the same classifiers used in the quantitative evaluation and edit by moving $z$ along the decision boundary of each classifier. See \cref{alg:trans} in \cref{app:inter-edit} for further details. Examples of images generated by using translations of $z$ with $x_T\sim\mathcal{N}(0,\mathbf{I})$, are shown in \cref{fig:edits}.

\begin{minipage}[t]{0.28\linewidth}
  \centering
  \vspace*{-6\baselineskip}
  \includegraphics[trim=0mm 0mm 54mm 0mm, clip, width=0.7\linewidth]{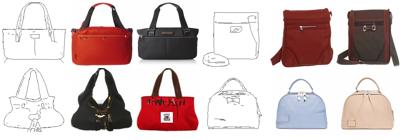}
  \includegraphics[trim=53mm 0mm 0mm 0mm, clip, width=0.7\linewidth]{imgs/examples/handbags.png}
  \captionof{figure}{Exemplar sketch-to-photo translations.}
  \label{fig:e2h}
\end{minipage}
\hfill
\begin{minipage}[b]{0.7\linewidth}
  \centering
  \vspace{\baselineskip}
  \captionof{table}{Sketch-to-photo translation: edges2handbags.}
  \label{tab:im2im}
  \begin{tabular}{@{}lcccc@{}}
    \toprule
    Model & FID $\downarrow$ & IS $\uparrow$ & LPIPS $\downarrow$ & MSE $\downarrow$ \\
    \midrule
    Pix2Pix \citep{pix2pix2017} & 74.8  & 4.24 & 0.356 & 0.209 \\
    DDIB \citep{su2023dualdiffusionimplicitbridges} & 186.84 & 2.04 & 0.869 & 1.05 \\
    SDEdit \citep{meng2022sdeditguidedimagesynthesis} & 26.5 & 3.58 & 0.271 & 0.510 \\
    Rectified Flow \citep{liu2022flowstraightfastlearning} & 25.3 & 2.80 & 0.241 & 0.088 \\
    I$^2$SB \citep{liu2023i2sbimagetoimageschrodingerbridge} & 7.43 & 3.40 & 0.244 & 0.191 \\
    DDBM (VE) \citep{zhou2023denoisingdiffusionbridgemodels} & 2.93 & 3.58 & 0.131 & 0.013 \\
    DDBM (VP) \citep{zhou2023denoisingdiffusionbridgemodels} & 1.83 & 3.73 & 0.142 & 0.040 \\\midrule
    \ours  & 3.28 & 3.59 & 0.359 & 0.209 \\
    \bottomrule
  \end{tabular}
\end{minipage}

\vspace*{\baselineskip}
\subsection{Multimodal framework}\label{sec:multi}

We demonstrate how \ours{} can be extended to handle multimodal tasks, specifically focusing on image-to-image translation. Inspired by Denoising Diffusion Bridge Models~\citep{zhou2023denoisingdiffusionbridgemodels}, we target the sketch-to-photo translation task: edges2handbags \citep{pix2pix2017}.

\paragraph{Model overview:}
To effectively reconstruct target ($x_0$) during translation, the model requires a sufficiently large \(|z|\). We train two separate \ours{}-512 models: one each for edge images (sketches) and handbags (photos).
Training terminates when the mean squared error (MSE) between inputs and samples generated using $z$ shows no further improvement (120K training iterations).
The two models learn independent latent spaces—$Z_{\text{sketch}}$ for edges and $Z_{\text{photo}}$ for handbags. We then learn the mapping between these two latent spaces: $\gamma: Z_{\text{sketch}} \to Z_{\text{photo}}$, where $\gamma$ is a mapping function parametrised by a multilayer perceptron (MLP).

\paragraph{Sketch-to-photo translation:}%
We follow this pipeline:
\begin{enumerate}[left=0pt,nosep,label=(\arabic*)]
    \item \textbf{Latent sampling}: We sample a latent variable $z_{\text{sketch}} \sim q_\phi(z | x_{\text{sketch}})$ from the model trained on sketches, where $x_{\text{sketch}}$ is the input sketch image.
    \item \textbf{Mapping}: We map sampled sketch latent $z_{\text{sketch}}$ into the photo latent space using the learned function $\gamma$, resulting in latent $z_{\text{photo}} = \gamma(z_{\text{sketch}})$.
    \item \textbf{Denoising}: Finally, we use a denoising process to generate the final photo image by sampling $p_\theta(x_T,z_{\text{photo}})$ from the photo model, where $x_T\sim \mathcal{N}(0,\mathbf{I})$.
\end{enumerate}

\paragraph{Results:}
Following \citet{zhou2023denoisingdiffusionbridgemodels}, we set $T=40$ and perform sketch-to-photo translations on the training set using our model and baselines (test-set results and rationale in \cref{app:im2im}).
We report FID scores, Inception Scores (IS), LPIPS \citep{zhang2018perceptual}, and Mean Squared Error (MSE) in \cref{tab:im2im}.
We perform competitively against existing approaches, showing effectiveness of the \ours{} framework for multimodal image translation. Crucially, \ours{} reduces the reliance on expensive joint training across domains.
Additionally, our framework supports unconditional generation of images in both domains (photos and sketches), reverse photo-to-sketch translation, and representation learning.
\cref{fig:e2h} provides examples of the edge-to-handbag translations.
Additional examples and discussion are provided in \Cref{app:im2im}.

\subsection{Ablations}
\label{sec:ablations}
We perform an ablation study to analyse the impact of various design choices on the model's abilities and performance. Specifically, we examine the use of discrete latents, conditioning denoising with \(z\), and the size of the latent variable $z$. Each of these choices plays a critical role in shaping the model's overall performance and capabilities.

Additionally, we demonstrate that \ours{} can be trained via finetuning a pretrained DDPM.
While finetuning leads to some loss in efficiency compared to training from scratch, it offers a faster training process, making it an attractive option when time or memory is a limiting factor.
Evaluation details and metrics used are as discussed in previous experiments.

\begin{wraptable}[9]{r}{0.65\textwidth}
  \centering
  \vspace{-1.5\baselineskip}
  \caption{Performance for discrete vs. continuous latents variables. Normal prior is fit over training data parameters. Comparison uses same number of training steps on CIFAR-10}
  \label{tab:contin}
  \small
  \begin{tabular}{@{}lcccccccc}
      \toprule
      \multirow{2}{*}{\shortstack{prior of $z$}} & \multirow{2}{*}{\shortstack{train\\iter.}} & \multirow{2}{*}{\shortstack{NLL\\(BPD)}} & \multirow{2}{*}{Acc} & \multicolumn{2}{c}{{FID@50K $\downarrow$}} &  \multicolumn{2}{c}{{IS@50K $\uparrow$}} \\
      \cmidrule(lr){5-6} \cmidrule(lr){7-8}
      & & & & \footnotesize{prior} & \footnotesize{$q_\phi(z|x_0)$} & \footnotesize{prior} & \footnotesize{$q_\phi(z|x_0)$} \\
      \midrule
      Normal  & 250K & 3.20 & 34.62 & 57.66 & 3.51 & 5.76 & 9.38 \\
      Bernoulli  &  250K & 3.18 & 39.50 & 2.62 & 2.79 & 9.64 & 9.77 \\
      \bottomrule
  \end{tabular}
  \vspace*{-\baselineskip}
\end{wraptable}

\paragraph{Discrete $z$:}
First, we explore alternatives to a discrete $z$, in the form of a Normal prior. We train \ours{}-16 on CIFAR-10 using two variants: discrete (Bernoulli prior enforced via Gumbel softmax \citep{jang2017categorical}) and continuous (Normal prior). \Cref{tab:contin} highlights the necessity of using discrete latents. The use of a continuous latent variable makes it infeasible to sample directly from the prior without auxiliary samplers that model the distribution of $z$. This is evidenced by the poor FID scores obtained when sampling directly from the prior, which, in the continuous version, we determine using the mean and standard deviation of the training data. Additionally, we evaluate a Gaussian Mixture Model (GMM) posterior, as proposed in DiffuseVAE \citep{pandey2022diffusevaeefficientcontrollablehighfidelity}. With 10, 20, 50, and 100 mixture components, we obtain FID@50K scores of 5.92, 6.95, 6.07 and 5.47, respectively, which is still substantially worse than those achieved by the discrete model. 
Furthermore, the quality of the learned representations declines, as reflected in the lower accuracy on downstream tasks. We attribute this to the latent space becoming more convoluted, making it more difficult for simple logistic classifiers to perform well.
We elaborate more on this ablation in \cref{app:contin}.

\begin{wraptable}[7]{r}{0.68\textwidth}
\vspace{-1.5\baselineskip}
  \centering
  \caption{Conditioning methods for $z$ with \ours{}-16 on CIFAR-10.}
  \label{tab:cond}
  \scalebox{0.93}{%
  \begin{tabular}{@{}lccccccc@{}}
    \toprule
    \multirow{2}{*}{method} & \multirow{2}{*}{\shortstack{train\\iter.}} & \multirow{2}{*}{\shortstack{NLL\\(BPD)}} & \multirow{2}{*}{Acc} &  \multicolumn{2}{c}{FID@50K $\downarrow$} & \multicolumn{2}{c}{IS@50K $\uparrow$}  \\
    \cmidrule(lr){5-6} \cmidrule(lr){7-8}
                            & & & & Bern & $q_\phi(z|x_0)$ & Bern & $q_\phi(z|x_0)$ \\
    \midrule
    Along with $t$ & 400K & 3.18 & 34.50  & 7.11 & 2.42 & 9.78 & 9.64 \\
    Cross-attention & 250K & 3.18 & 39.50 & 2.62 & 2.79 & 9.64 & 9.77 \\
    \bottomrule
  \end{tabular}}
\end{wraptable}
\paragraph{Conditioning via cross-attention:}
We investigate different strategies for incorporating $z$ into the denoising network. The denoising, unconditional UNets consist of ResNet blocks, up/downsampling blocks, and self-attention blocks. These UNets are conditioned on the timestep $t$ by passing $t$ to each ResNet block.
A straightforward way to incorporate $z$ is to provide it alongside the timestep $t$ to each ResNet block \citep{Preechakul_2022_CVPR,pmlr-v202-wang23ah}, e.g., by concatenating their embeddings and using the result in place of the standard timestep embedding \citep{Preechakul_2022_CVPR}.
However, we find that incorporating $z$ through cross-attention results in better performance.
\cref{fig:cond} highlights the architectural differences between the two approaches.

We replace selected self-attention blocks in the U-Net with cross-attention, enabling better attention over $z$. This improves robustness to $z$ values not seen during training and leads to better learned representations---reflected in both lower FID scores when sampling from the Bernoulli prior and higher accuracy on downstream tasks (\cref{tab:cond}).
Note that while our experiments here are with the U-Net model to ensure fair comparison with other DAs, this cross-attention strategy naturally extends to more recent advances such as Diffusion Transformers \citep{Peebles2022DiT}. We include results with DiT architectures in \Cref{app:dit}, noting that the relative gains from including \(z\) also hold there.

\paragraph{Small size of $z$:}
For an autoencoder to accurately reconstruct $x_0$ from $z$, the latent variable $z$ must be sufficiently large. However, the opposite holds true for effective generation: a smaller latent space tends to yield better generative performance \citep[e.g.,][]{yao2025reconstruction}.
Furthermore, a small $|z|$ is sufficient for nearly all use cases. The only exception occurs when reconstruction from $z$ is required, such as in image-to-image translation tasks where Mean Squared Error (MSE) is of concern. Even for image manipulation, a small $|z|$ suffices, as the additional information that a larger $|z|$ could provide is already encoded in the intermediate $x_t$, which is accessible (unlike in image-to-image translation).

We observe that for larger $|z|$, sampling from the Bernoulli prior becomes less effective. To address this, we explore several strategies for sampling $z$ during inference---a critical component for high-quality generation, as evidenced in prior work on DAs.

\begin{wraptable}[12]{r}{0.5\textwidth}
  \centering
  \caption{FID@50K scores of sampling strategies for \ours{} trained on CIFAR-10.}
  \label{tab:sampler}
  \begin{tabular}{c@{\hspace*{3ex}}l@{\hspace*{3ex}}rrrr}
      \toprule
      \multirow{2}{*}{$T$} & \multirow{2}{*}{$z \sim \bullet$} & \multicolumn{4}{c}{$\lvert z \rvert$} \\ \cmidrule(lr){3-6}
      & &  16 & 32 & 64 & 128 \\
      \midrule
      \multirow{3}{*}{10} & $q_\phi(z|x_0)$ & 9.70 & 8.45 & 7.14 & 7.34 \\
      &$\text{Bernoulli}$ & 10.03 &  8.09 & 13.78 & 20.57 \\
      & $\text{PixelSNAIL}$ &  9.75 & 9.08 & 9.45 & 12.73 \\
      \midrule
      \multirow{3}{*}{100} & $q_\phi(z|x_0)$ &  2.79  & 2.76 & 2.51 & 2.45 \\
      & $\text{Bernoulli}$ & 2.62 & 3.11 & 7.14 & 15.09 \\
      & $\text{PixelSNAIL}$ & 2.65 & 2.87 & 3.94 & 7.54 \\
      \bottomrule
  \end{tabular}
  \vspace*{-1.5\baselineskip}
\end{wraptable}
We consider the following three methods:
\begin{enumerate}[left=0pt,nosep,label=(\arabic*)]
\item \textbf{Sampling $z$ from data:} For reference, we compute FID scores for $z \sim q_\phi(z \mid x_0)$, where $x_0 \sim \mathcal{D}$ is taken from data. We denote this strategy as $z\sim q_\phi(z|x_0)$.
\item \textbf{Bernoulli Prior:} We sample each latent component independently as $z_i \sim \text{Bernoulli}(p=0.5)$.
\item \textbf{Autoregressive Prior (PixelSNAIL):} Inspired by prior work on discrete latent models \citep{oord2018neuraldiscreterepresentationlearning,razavi2019generatingdiversehighfidelityimages}, we fit a PixelSNAIL model \citep{chen2017pixelsnailimprovedautoregressivegenerative} over latent codes to enable sampling. We refer to this sampling method as $z \sim \text{PixelSNAIL}$.
\end{enumerate}

Larger PixelSNAIL models closely match the posterior, or even memorise the data, achieving FID scores near those for latents from data.
To ensure fair comparison, our models are limited to \(<\)600K parameters, using grid search to choose hyperparameters providing an optimal balance between performance and model size.

FID scores for all strategies are shown in \cref{tab:sampler}.
Sampling from PixelSNAIL generally yields better results in higher-dimensional settings.
In lower-dimensional settings, the model better leverages the prior, and direct sampling yields strong performance without auxiliary samplers.
The degradation of direct sampling at larger $|z|$ follows directly from the coverage argument: for small $|z|$, a large fraction of the $2^{|z|}$ possible codes are encountered during training, reducing the number of unseen configurations at inference.
This coverage breaks down as $|z|$ grows.
We view this as a positive design principle: keeping $|z|$ small enables sampler-free inference.
Therefore, we adopt low-dimensional $z$, optimising for direct sampling.

\paragraph{Finetuning:}
All models presented thus far were trained from scratch. Here, we study how finetuning strategies affect performance. Since time and computational resources may be limited, one may opt to finetune a pretrained DM to accelerate training. We explore how this choice affects discussed capabilities.

We consider three training strategies: 1) training from scratch, 2) finetuning all parameters, 3) finetuning only the newly added parameters (specifically, those related to the cross-attention mechanism) using a pretrained DDPM.
Note that the pretrained DDPM is unconditional and from prior experiments. Results are in \cref{tab:fine}.

\begin{wraptable}[8]{r}{0.66\textwidth}
\vspace{-1.5\baselineskip}
  \centering
  \caption{\ours{}-64: finetuning strategies on CelebA-64.}
  \label{tab:fine}
  \scalebox{0.93}{%
  \begin{tabular}{@{}lccccccccc@{}}
      \toprule
      \multirow{2}{*}{\shortstack{Finetuning}} & \multirow{2}{*}{\shortstack{train\\iter.}} & \multirow{2}{*}{\shortstack{NLL\\(BPD)}} & \multirow{2}{*}{AUROC} & \multicolumn{2}{c}{FID@50K $\downarrow$} & \multicolumn{2}{c}{IS@50K $\uparrow$} \\
      \cmidrule(lr){5-6} \cmidrule(lr){7-8}
      & & & & \footnotesize{T=10} & \footnotesize{T=100} & \footnotesize{T=10} & \footnotesize{T=100} \\
      \midrule
      None & 300K & 2.61 & 86.4 & 14.43 & 2.93 & 3.28 & 3.27 \\
      All params & 100K & 2.65 & 79.7 & 17.75 & 3.29 & 3.26 & 2.93\\
      New params & 100K & 2.66 & 82.5 & 18.91 & 2.34 & 3.14 & 3.25 \\
      \bottomrule
  \end{tabular}}
\end{wraptable}

All models were able to learn effective representations, as demonstrated by their performance on downstream tasks (AUROC). However, the finetuned models did not perform as well in generation tasks with fewer denoising steps, as evidenced by the FID scores for $T=10$.
Overall, all models perform well and are suitable for different use cases, depending on the specific trade-offs between training time, resource requirements, and generation efficiency.
In \Cref{app:finetune}, we present experiments on finetuning larger-scale \ours{} using pretrained models from Hugging Face.

\section{Conclusion}
We presented \ours{}, a model inspired by the connection between diffusion autoencoders and diffusion models with a learnable forward process and designed to both learn efficiently and capture effective representations. 
Through targeted experimentation, we demonstrate that \ours{} is capable of generating high-quality samples with fewer denoising steps, while simultaneously learning meaningful representations.
Importantly, \ours{} achieves these results without additional loss terms or latent-variable constraints, relying solely on a binary latent variable $z$ injected via cross-attention. With a small latent, $z$ can be sampled directly from a simple prior; larger latents, which favour reconstruction-heavy uses such as image-to-image translation, trade this simplicity for a learned prior.
Finally, we extend \ours{} to a multimodal framework and successfully apply it to an image-to-image translation task, showcasing its versatility and effectiveness.
Our findings suggest that input-dependent latent variables offer a promising alternative to traditional diffusion modelling when training iterations and sampling steps are the bottleneck.

\section*{Acknowledgements}

We acknowledge support of the UKRI AI Hub in Generative Models [EP/Y028805/1].
ESW acknowledges support from the CIFAR Learning in Machines and Brains Programme.



\section*{Impact Statement}

This paper presents work whose goal is to advance the field of Machine
Learning. There are many potential societal consequences of our work, none
which we feel must be specifically highlighted here.

\bibliography{main}
\bibliographystyle{tmlr}

\newpage
\appendix

\crefalias{section}{appendix}
\crefalias{subsection}{appendix}

\crefname{appendix}{appendix}{appendices}
\Crefname{appendix}{Appendix}{Appendices}

\section{Appendix}

\subsection{\ours{} in relation to prior work}
In \Cref{tab:comparison}, we show how \ours{} relates to prior work on diffusion autoencoders.
\Cref{alg:train,alg:sample} outline the training and sampling procedures of \ours{}, detailing how the latent variable $z$ is incorporated and highlighting differences from the original DDPM approach \citep{ho2020denoisingdiffusionprobabilisticmodels}.

\begin{table}[htb!]
\centering
\caption{Overview of model features related to training, representations, and evaluation. Diff. loss only -- optimisation solely using the diffusion loss. Repr. qual. — quantitative evaluation of representation quality, e.g., via downstream task performance metrics. $^*$ For small latent dimensionality $|z|$.}
\label{tab:comparison}
\begin{adjustbox}{width=\linewidth}
\begin{tabular}{@{}lccccccccc}
\toprule
& \multicolumn{2}{c}{\textbf{Training}}
& \multicolumn{2}{c}{\textbf{Representations}}
& \multicolumn{5}{c}{\textbf{Evaluation}} \\
\cmidrule(lr){2-3} \cmidrule(lr){4-5} \cmidrule(lr){6-10}
& \multirow{2}{*}{E2E}
& \multirow{2}{*}{\shortstack[c]{Diff. loss\\ only}}
& \multirow{2}{*}{Discrete}
& \multirow{2}{*}{\shortstack[c]{No aux \\ sampler}}
& \multirow{2}{*}{\shortstack[c]{HQ\\samples}}
& \multirow{2}{*}{Edits}
& \multirow{2}{*}{\shortstack[c]{Repr. \\ qual.}}
& \multirow{2}{*}{\shortstack[c]{Multi- \\ modal}}
& \multirow{2}{*}{\shortstack[c]{Fine-\\tuning}} \\
& & & & & & & & & \\
\midrule
DiffAE \citep{Preechakul_2022_CVPR} & \ding{51} & \ding{51} & \ding{55} & \ding{55} & \ding{51} & \ding{51} & \ding{51} & \ding{55} & \ding{55} \\
InfoDiff \citep{pmlr-v202-wang23ah}& \ding{51} & \ding{55} & \ding{51} & \ding{55} & \ding{51} & \ding{51} & \ding{51} & \ding{55} & \ding{55} \\
DiffuseVAE \citep{pandey2022diffusevaeefficientcontrollablehighfidelity} & \ding{55} & \ding{55} & \ding{55} & \ding{55} & \ding{51} & \ding{51} & \ding{55} & \ding{55} & \ding{55} \\
\ours{} & \ding{51} & \ding{51} & \ding{51} & \ding{51}$^*$ & \ding{51} & \ding{51} & \ding{51} & \ding{51} & \ding{51} \\
\bottomrule
\end{tabular}
\end{adjustbox}
\end{table}

\begin{minipage}[t]{0.48\linewidth}
\centering
\begin{algorithm}[H]
\caption{\ours{} training}
\label{alg:train}
\begin{algorithmic}
  \REPEAT
  \STATE Sample $x_0 \sim q(x_0)$, $t \sim \mathcal{U}(\{1, \dots, T\})$, $\epsilon \sim \mathcal{N}(0, \mathbf{I})$
  \STATE Compute noisy input $x_t \gets \sqrt{\bar{\alpha}_t} x_0 + \sqrt{1 - \bar{\alpha}_t} \pmb{\epsilon}$
  \STATE Extract relaxed code $z$ from $x_0$ via encoder parametrised by $\varphi$
  \STATE Take a gradient step on
  $\nabla_{\theta,\varphi} \left\| \epsilon - \epsilon_\theta(x_t, t, z) \right\|^2$
  \UNTIL{convergence}
\end{algorithmic}
\end{algorithm}
\end{minipage}
\hfill
\begin{minipage}[t]{0.48\linewidth}
\centering
\begin{algorithm}[H]
\caption{\ours{} sampling}
\label{alg:sample}
\begin{algorithmic}
  \STATE Sample $\hat{x}_T \sim \mathcal{N}(0, \mathbf{I})$ and $z$ such that $z_i\sim \text{Bernoulli}(p=0.5)$
  \FOR{$t \gets T$ to $1$}
  \IF{$t > 1$}
  \STATE Sample $v \sim \mathcal{N}(0, \mathbf{I})$
  \ELSE
  \STATE Set $v \gets 0$
  \ENDIF
  \STATE $\hat{x}_{t-1} \gets \frac{1}{\sqrt{\alpha_t}} \left( \hat{x}_t - \frac{1 - \alpha_t}{\sqrt{1 - \bar{\alpha}_t}} {\epsilon}_\theta(\hat{x}_t, t, z) \right) + \sigma_t v$
  \ENDFOR
  \STATE \textbf{return} $\hat{x}_0$
\end{algorithmic}
\end{algorithm}
\end{minipage}

\subsection{Evidence That $z$ Is Not Ignored} \label{app:proof}

\begin{lemma}[Conditional law of total variance]
For random variables $X$, $Y$, and $Z$,
$$
\operatorname{Var}(Y\mid X)
=
\mathbb{E}[\operatorname{Var}(Y\mid X,Z)\mid X]
+
\operatorname{Var}(\mathbb{E}[Y\mid X,Z]\mid X).
$$
\end{lemma}

\begin{proof}
Starting from the definition of conditional variance, we have
$$\operatorname{Var}(Y\mid X) = \mathbb{E}[Y^2\mid X] - \mathbb{E}[Y\mid X]^2.$$
By the law of total expectation,
$$\mathbb{E}[Y^2\mid X]=\mathbb{E}\!\left[\mathbb{E}[Y^2\mid X,Z]\mid X\right],$$
and using the definition of conditional variance $\operatorname{Var}(Y\mid X,Z)$:
$$\mathbb{E}[Y^2\mid X,Z]=\operatorname{Var}(Y\mid X,Z)+\mathbb{E}[Y\mid X,Z]^2.$$

Therefore,
$$
\mathbb{E}[Y^2\mid X]
=
\mathbb{E}[\operatorname{Var}(Y\mid X,Z)\mid X]
+
\mathbb{E}\!\left[\mathbb{E}[Y\mid X,Z]^2\mid X\right].
$$

Substituting into the expression for $\operatorname{Var}(Y\mid X)$ yields
$$
\operatorname{Var}(Y\mid X)
=
\mathbb{E}[\operatorname{Var}(Y\mid X,Z)\mid X]
+
\mathbb{E}\!\left[\mathbb{E}[Y\mid X,Z]^2\mid X\right]
-
\mathbb{E}[Y\mid X]^2.
$$

By the law of total expectation,
$$
\mathbb{E}[Y\mid X]
=
\mathbb{E}\!\left[\mathbb{E}[Y\mid X,Z]\mid X\right].
$$

Applying the definition of conditional variance to the random variable
$\mathbb{E}[Y\mid X,Z]$ conditioned on $X$ gives
$$
\operatorname{Var}(\mathbb{E}[Y\mid X,Z]\mid X)
=
\mathbb{E}\!\left[\mathbb{E}[Y\mid X,Z]^2\mid X\right]
-
\mathbb{E}[Y\mid X]^2.
$$

Substituting this identity into the previous expression yields
$$
\operatorname{Var}(Y\mid X)
=
\mathbb{E}[\operatorname{Var}(Y\mid X,Z)\mid X]
+
\operatorname{Var}(\mathbb{E}[Y\mid X,Z]\mid X),
$$
which proves the claim.
\end{proof}

We investigate the conditions under which the latent variable is utilised. To this end, we compare the optimal denoising estimators with and without the latent code $z$, corresponding to DMZ and DDPM, respectively, with minimum mean-squared errors:
$$\mathcal{L}_{\text{DMZ}} = \mathbb{E}[\operatorname{Var}(\epsilon \mid x_t,z)]
\qquad
\mathcal{L}_{\text{DDPM}} = \mathbb{E}[\operatorname{Var}(\epsilon \mid x_t)].$$

By the law of total variance applied conditionally on $x_t$, we have
$$\operatorname{Var}(\epsilon \mid x_t)=\mathbb{E}[\operatorname{Var}(\epsilon \mid x_t, z)\mid x_t]+\operatorname{Var}(\mathbb{E}[\epsilon \mid x_t, z]\mid x_t),$$ which yields
$$
\mathcal{L}_{\text{DDPM}} =\mathcal{L}_{\text{DMZ}} +\mathbb{E}\left[\operatorname{Var}\left(\mathbb{E}[\epsilon \mid x_t,z] \mid x_t\right)\right].$$

The second term is non-negative and is strictly positive whenever $z$ contains information about the clean sample that is not already present in $x_t$, i.e., when $I(X;Z \mid X_t) > 0$. Therefore,
$$
\mathcal{L}_{\text{DMZ}} \le \mathcal{L}_{\text{DDPM}}.
$$
Consequently, a network that ignores $z$ cannot achieve the minimum loss unless $Z \perp X \mid X_t$. Hence, optimization is pushed to exploit $z$ whenever it provides additional information beyond $x_t$.

\paragraph{Mispaired-$z$ control.}\label{app:mispaired}
To test whether the denoiser relies on the image-specific code, we reconstruct 10K CIFAR-10 images with a DMZ-16 model ($T=100$, $x_{\text{pred}}\sim p_\theta(x_0\mid x_T,z)$) and measure $\mathrm{MSE}(x_0,x_{\text{pred}})$ under three settings:
\begin{enumerate}[left=1ex,nosep]
    \item the paired code $z\sim q_\phi(z\mid x_0)$,
    \item the code of a randomly selected different image, $z\sim q_\phi(z\mid x_0')$, and
    \item a prior sample $z\sim\mathrm{Bernoulli}(0.5)$.
\end{enumerate}
The MSE is $0.40\pm0.21$, $0.45\pm0.20$ and $0.46\pm0.22$, respectively, so reconstruction is worse when the code does not belong to the image, showing that the network uses $z$.
For generation, permuting codes across images does not change FID, since codes are drawn from the dataset in both cases.
The informative comparison is dataset codes versus prior samples (\cref{tab:sampler}): FID $2.79$ versus $2.62$ at $T=100$ and $9.70$ versus $10.03$ at $T=10$.
The small difference is consistent with the near-uniform aggregate posterior ($\mathrm{KL}(q_\phi(z)\,\|\,p(z))=0.0006$ nats).

\paragraph{Not required vs uninformative $z$.}
The standard objective can be optimised without $z$, so it places no pressure on $z$ to be informative.
Under a variational objective, the KL term penalises encoded information, so ignoring $z$ is a cheap solution and posterior collapse is likely (\cref{app:kldiv}); this is why we omit prior regularisation. Without the KL term, the only force acting on $z$ is the denoising loss, which rewards information that reduces $\mathbb{E}[\mathrm{Var}(\epsilon\mid x_t,z)]$, so $z$ is informative \emph{when useful}, without being required. Empirically, $z$ is used by the denoiser (\cref{fig:mi} and mispaired control above) and supports classification (\cref{tab:acc}).

\subsection{Effect of Regularisation Against Non-informative Prior} \label{app:kldiv}

\ours{} learns an aggregate posterior that aligns with the prior, enabling sampling of $z$ during inference. However, this alignment is not achieved for larger values of $z$. A reasonable solution is to apply regularisation techniques, such as a KL regularisation, to better align the distributions. \Cref{tab:kl-regularizer-results} compares the performance of \ours{} with and without KL regularisation.

\begin{wraptable}[8]{r}{0.4\textwidth}
\centering
\small
\vspace{-2\baselineskip}
\caption{Performance on CIFAR-10 with and without KL regularisation}
\label{tab:kl-regularizer-results}
\begin{tabular}{cccccc}
\toprule
\multirow{2}{*}{$|z|$} & \multirow{2}{*}{KL reg.} & \multirow{2}{*}{Acc} & \multicolumn{2}{c}{FID@10K} \\
& & & Bernoulli & $q_\phi(z|x_0)$ \\
\midrule
16 & \ding{55}  & 39.5 & 4.79 & 4.56 \\
16 & \ding{51} & 35.8 & 5.05 & 5.23 \\
64 & \ding{55}  & 45.6 & 9.33 & 4.61 \\
64 & \ding{51} & 34.9 & 5.40 & 6.05 \\
\bottomrule
\end{tabular}
\end{wraptable}

As expected, with regularisation, the models learn to better align the posterior with the prior. However, this leads to less meaningful representations that do not improve the generative performance of the model.
While good samples can be generated using the Bernoulli prior, the generative quality is weaker compared to when $|z|$ is restricted to $16$ to achieve prior-posterior alignment. Moreover, regularisation negatively impacts the generative performance of the model, even for $|z|=16$, as it interferes with the denoiser's objective.

\subsection{Training and sampling-step efficiency under budget constraints} \label{app:efficiency}

Our efficiency claims concern two distinct notions.
\emph{Training efficiency} refers to the number of training iterations required to converge.
\emph{Sampling-step efficiency} refers to achieving better generation quality (FID) with fewer denoising steps $T$, particularly at small $T$.
We do \emph{not} claim lower per-step deployment cost or wall-clock inference time.

\paragraph{Sampling-step efficiency.}
\cref{tab:app_sampling} compares \ours{}-16 with DDPM baselines on CIFAR-10.  The 51M-parameter DDPM is the baseline reported in the main text and differs from \ours{} only in its cross-attention components. The 64M and 76M DDPMs are closely capacity-matched baselines, since no single DDPM configuration exactly matches the size of \ours{}-16 (68M). 

We experimented with varying model depth and width. However, increasing the width from 128 to 160 (the smallest feasible increase without changing other hyperparameters) resulted in an 80M-parameter model, which deviates substantially from the target size. In contrast, using 4 and 5 residual blocks yielded 64M and 76M parameters, respectively. We therefore selected the latter configurations for a fairer comparison. The remaining hyperparameters are listed in Table~\ref{tab:hparams}, with implementation details provided in the shared codebase.

\begin{table}[h]
\centering
\caption{FID and sampling time on CIFAR-10 for different numbers of denoising steps $T$. \ours{}-16 achieves substantially better FID at low $T$, even compared with the larger 76M DDPM. Time is reported in seconds for a single batch of 64.}
\label{tab:app_sampling}
\small
\begin{tabular}{lcccccc}
\toprule
Model & U-Net & Metric & $T{=}10$ & $T{=}20$ & $T{=}50$ & $T{=}100$ \\
\midrule
DDPM & 51M & FID  & 21.25 & 7.21 & 3.18 & 2.56 \\
     &     & Time & 0.76$\pm$0.00 & 1.46$\pm$0.00 & 3.74$\pm$0.00 & 7.16$\pm$0.01 \\
\midrule
DDPM & 64M & FID  & 21.49 & 7.02 & 3.05 & 2.60 \\
     &     & Time & 0.89$\pm$0.00 & 1.79$\pm$0.00 & 4.50$\pm$0.01 & 9.02$\pm$0.01 \\
\midrule
DDPM & 76M & FID  & 19.23 & 5.96 & \textbf{2.98} & \textbf{2.55} \\
     &     & Time & 1.08$\pm$0.00 & 2.17$\pm$0.01 & 5.42$\pm$0.01 & 10.91$\pm$0.01 \\
\midrule
\ours{}-16 & 68M & FID  & \textbf{9.78} & \textbf{4.79} & 3.04 & 2.79 \\
     &     & Time & 1.15$\pm$0.00 & 2.30$\pm$0.00 & 5.76$\pm$0.00 & 11.09$\pm$0.01 \\
\bottomrule
\end{tabular}
\end{table}

At $T{=}10$, \ours{}-16 achieves an FID of 9.78 versus 19.23 for the larger 76M DDPM, and at $T{=}20$ it achieves 4.79 versus 5.96.
The improvement at low $T$ is therefore not explained by model capacity.
For $T \geq 50$ the models perform comparably, with the larger DDPMs slightly ahead at $T{=}100$.
The runtimes show that \ours{} incurs a per-step overhead, primarily due to the additional cross-attention blocks.
We do not claim an optimised wall-clock inference advantage, as runtime depends substantially on hardware, parallelism, and implementation-level optimisations, which we do not optimise for here.

\paragraph{Training efficiency.}
\cref{tab:app_training} reports training cost. GFLOPs are measured with the \texttt{fvcore} library\footnote{\url{https://github.com/facebookresearch/fvcore}}.
\ours{}-16 converges in 250K iterations, compared with 450K--550K for the DDPM baselines.

\begin{table}[h]
\centering
\caption{Training cost on CIFAR-10. \ours{} is slower per iteration, but it converges in far fewer iterations.}
\label{tab:app_training}
\small
\begin{tabular}{lcccc}
\toprule
Model & U-Net size & Training iter. & Time / 10K iter. & GFLOPs \\
\midrule
DDPM & 51M & 550K & $\sim$38 min & 8.12 \\
DDPM & 64M & 550K & $\sim$44 min & 10.13 \\
DDPM & 76M & 450K & $\sim$54 min & 12.13 \\
\ours{}-16 & 68M & 250K & $\sim$56 min & 10.51 \\
\bottomrule
\end{tabular}
\vspace{-0\baselineskip}
\end{table}

While \ours{} has a higher training time per iteration, it converges in far fewer iterations.
This is also visible in the CelebA training curves in \cref{fig:train}, while the FID results in \cref{tab:app_sampling} further support the improved generation quality at low $T$.

The gain also persists at longer training: at 550K iterations, \ours{} reaches 3.16 bits per dimension (BPD) on CIFAR-10 versus 3.18 for DDPM, and it already reaches 3.18 BPD at 250K iterations.
\ours{} therefore provides both faster convergence and a slightly improved final likelihood.

\Cref{fig:pareto} shows that \ours{}-16 lies on the FID--time frontier at the lowest budgets ($\lesssim 2.5$\,s per batch, $T=10,20$), but that DDPMs dominate it from roughly $3$\,s onwards. Our claim is therefore limited to sampling-step efficiency at small step budgets, not to wall-clock efficiency in general.

\begin{figure}[h]
\centering
\begin{tikzpicture}
\begin{axis}[
    width=0.85\linewidth,
    height=6cm,
    xlabel={Sampling time (s)},
    ylabel={FID@50K},
    ymode=normal,
    ymin=0, ymax=24,
    xmin=0, xmax=12,
    grid=major,
    grid style={gray!20},
    tick label style={font=\small},
    label style={font=\small},
    legend style={
        font=\small,
        at={(0.98,0.98)},
        anchor=north east,
        draw=gray!30,
        fill=white,
        rounded corners=2pt
    },
    line width=1.1pt,
    mark size=2.8pt,
]
\definecolor{settwo1}{HTML}{66C2A5}
\definecolor{settwo2}{HTML}{FC8D62}
\definecolor{settwo3}{HTML}{8DA0CB}
\definecolor{settwo4}{HTML}{E78AC3}

\addplot[color=settwo1, mark=o]
    coordinates {(0.76,21.25)(1.46,7.21)(3.74,3.18)(7.16,2.56)};
\addlegendentry{DDPM 51M}

\addplot[color=settwo4, mark=square*]
    coordinates {(0.89,21.49)(1.79,7.02)(4.50,3.05)(9.02,2.60)};
\addlegendentry{DDPM 64M}

\addplot[color=settwo3, mark=triangle*]
    coordinates {(1.08,19.23)(2.17,5.96)(5.42,2.98)(10.91,2.55)};
\addlegendentry{DDPM 76M}

\addplot[color=settwo2, mark=*, very thick, mark size=3.2pt]
    coordinates {(1.15,9.78)(2.30,4.79)(5.76,3.04)(11.09,2.79)};
\addlegendentry{\ours{}-16 (68M)}

\end{axis}
\end{tikzpicture}
\caption{FID versus sampling time on CIFAR-10 for different sampling steps ($T=10,20,50,100$).}
\label{fig:pareto}
\vspace{\baselineskip}
\end{figure}

\subsection{\ours{} with DDIM sampling}\label{app:ddim}

Our baseline DDPM is incompatible with DDIM sampling due to its use of the input perturbation technique (see Section 6.2 in \citep{ning2023inputperturbationreducesexposure}) within the DDPM framework\footnote{\url{https://github.com/openai/guided-diffusion}}.
To evaluate the effect of DDIM on DMZ, we trained models without the input perturbation technique.
The results in \Cref{tab:ddim} indicate that DDIM sampling and DMZ are orthogonal methods, each independently improving generative quality.
Note that the DDPM without input perturbation performs worse than our baseline DDPM (see \Cref{tab:fid}), which is why we opt to use the latter.

\begin{table}[h]
    \centering
    \small
    \vspace{-\baselineskip}
    \caption{Evaluation of the effect of DDIM sampling on \ours{} measured via FID@10K.}
    \label{tab:ddim}
    \begin{tabular}{lccccccccc}
        \toprule
        \multirow{3}{*}{Model} & \multirow{3}{*}{\shortstack[]{DDIM\\sampling}}  & \multicolumn{4}{c}{CIFAR-10} & \multicolumn{4}{c}{CelebA-64} \\
        \cmidrule(lr){3-6} \cmidrule(lr){7-10}
        & & 10 & 20 & 50 & 100 & 10 & 20 & 50 & 100 \\
        \midrule
        \multirow{2}{*}{\ours{}} &  \ding{55}  & 15.44 & 8.88 & 5.99 & 5.11 & 28.62 & 13.73 & 6.72 & 4.87   \\
        &  \ding{51} & 12.51 & 8.55 & 6.19 & 6.08 & 20.95 & 12.42 & 6.51 & 4.52   \\
        \midrule
        \multirow{2}{*}{DDPM}    &  \ding{55}  & 22.90 & 11.93 & 6.80 & 5.53 & 73.14 & 29.35 & 11.39 & 8.43   \\
            & \ding{51} & 13.17 & 8.21 & 5.54 & 5.29 & 20.34 & 14.91 & 10.31 & 9.41    \\
        \bottomrule
    \end{tabular}
    \vspace{-\baselineskip}
\end{table}

\subsection{Interpolation formulation and examples}
\label{app:inter-edit}

\begin{figure}[!ht]
    \vspace{-0.5\baselineskip}
    \centering
    \begin{subfigure}[b]{0.48\textwidth}
        \includegraphics[trim=0 2 0 1, clip, width=\linewidth]{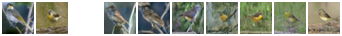}
        \includegraphics[trim=0 2 0 1, clip, width=\linewidth]{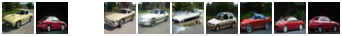}
        \includegraphics[trim=0 2 0 1, clip, width=\linewidth]{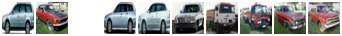}
        \includegraphics[trim=0 2 0 1, clip, width=\linewidth]{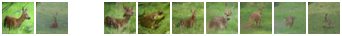}
        \caption{\ours{}-128 for CIFAR-10}
    \end{subfigure}
    \hfill
    \begin{subfigure}[b]{0.48\textwidth}
        \includegraphics[trim=0 3 0 3, clip, width=\linewidth]{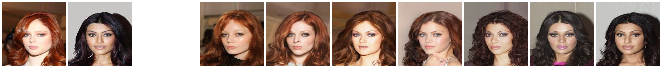}
        \includegraphics[trim=0 3 0 3, clip, width=\linewidth]{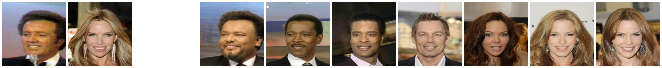}
        \includegraphics[trim=0 3 0 3, clip, width=\linewidth]{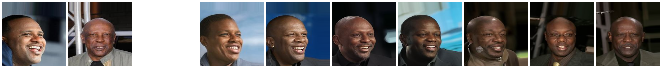}
        \includegraphics[trim=0 3 0 3, clip, width=\linewidth]{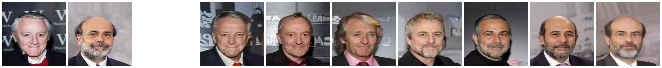}
        \caption{\ours{}-256 for CelebA-64}
    \end{subfigure}
    \caption{Examples of discrete interpolations between codes $z^a$ and $z^b$, where $z^a \sim q_\phi(z|x_0^a), z^b\sim q_\phi(z|x_0^b)$, $x_0^a,x_0^b\sim\mathcal{D}$, and $x_T\sim N(0,\mathbf{I})$. Original images $x_0^a,x_0^b$ are shown in the first two columns.}
    \label{fig:inter-disc}
    \vspace{\baselineskip}
\end{figure}

\begin{figure}[t]
    \centering
    \begin{subfigure}[b]{0.4\textwidth}
         \includegraphics[width=\linewidth]{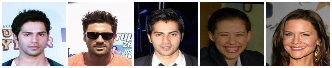}
        \caption{Input images $x_0\sim \mathcal{D}$}
         \vspace*{1mm}
    \end{subfigure}
    \\
    \begin{subfigure}[b]{0.45\textwidth}
         \includegraphics[width=\linewidth]{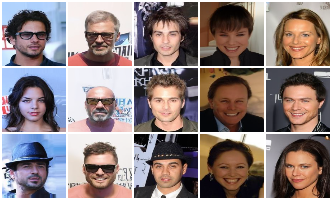}
        \caption{$z\sim q_\varphi(z|x_0)$ with $x_T\sim \mathcal{N}(0,\mathbf{I})$}
        \vspace*{1mm}
    \end{subfigure}
    \hfill
    \begin{subfigure}[b]{0.45\textwidth}
        \includegraphics[width=\linewidth]{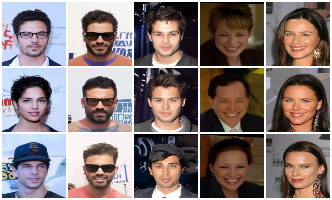}
        \caption{$z\sim q_\varphi(z|x_0)$ with $ x_t\sim q(x_t|x_0)$ and $t=90$}
    \end{subfigure}
    \caption{Examples of classifier-based edits for $T=100$ on CelebA-64 using \ours{}-256. For the first image $x_0$, we change following CelebA attributes: glasses, male, hat; for the second: gray hair, bald, smile; the third: bangs, blond hair, hat; the fourth: bangs, male, earrings; the fifth: blond hair, male, smile.}
    \label{fig:inter-class}
    \vspace{0.5\baselineskip}
\end{figure}

\Cref{alg:inter} describes how discrete interpolations between two latent vectors are performed by flipping the disagreeing bits one at a time in random order. For visualisations, we take latent codes equally spaced along the interpolation trajectory.
\Cref{alg:trans} details translations across the classifier’s decision boundary.
Examples of both are shown in \Cref{fig:inter-disc} and \Cref{fig:inter-class}.

\begin{minipage}[t]{0.48\linewidth}
\vspace{-\baselineskip}
\centering
\begin{algorithm}[H]
\caption{Discrete interpolation: $z_{\text{source}} \to z_{\text{target}}$}
\label{alg:inter}
\begin{algorithmic}
  \STATE \textbf{Input:} $z_{\text{source}}, z_{\text{target}} \in \{0,1\}^n$
  \STATE \textbf{Output:} Interpolation $\{z^{(i)}\}_{i=0}^{k},\, z^{(i)} \in \{0,1\}^n$, $z^{(0)}=z_{\text{source}}$, $z^{(k)}=z_{\text{target}}$
  \STATE $\mathcal{I}=\{j\mid {z_{\text{source}}}_j\neq{z_{\text{target}}}_j\}$
  \COMMENT{disagreement indices}
  \STATE Let $k=|\mathcal{I}|$ and $j_1,\dots,j_k$ be a random ordering of $\mathcal{I}$
  \STATE $z^{(0)} \gets z_{\text{source}}$\\
  \FOR{$i \gets 1$ to $k$}
  \STATE $z^{(i)} \gets z^{(i-1)}$
  \STATE $z^{(i)}_{j_i} \gets 1 - z^{(i)}_{j_i}$ \COMMENT{flip bit}
  \ENDFOR
  \STATE \textbf{return} $\{z^{(i)}\}_{i=0}^{k}$
\end{algorithmic}
\end{algorithm}
\end{minipage}
\hfill
\begin{minipage}[t]{0.48\linewidth}
\vspace{-\baselineskip}
\centering
\begin{algorithm}[H]
\caption{Translating \(z\) across binary decision boundary}
\label{alg:trans}
\begin{algorithmic}
  \STATE \textbf{Input:} $z \in \{0,1\}^n$, classifier params $W \in \mathbb{R}^{n \times 2},\, b \in \mathbb{R}^2$, step sizes with directions $\delta_i\in\mathbb{R}, i=1,\dots,k$
  \STATE \textbf{Output:} Interpolation $\{z^{(i)}\}_{i=0}^{k},\, z^{(i)} \in \{0,1\}^n$
  \STATE Let $w_1 \gets W[:, 0]$, $w_2 \gets W[:, 1]$ \COMMENT{class weights}
  \STATE $\mathbf{u} \gets w_1 - w_2$ \COMMENT{normal vector to decision boundary}
  \STATE $\mathbf{v} \gets \frac{\mathbf{u}^\top z + b_1 - b_2}{\|\mathbf{u}\|^2} \cdot \mathbf{u}$ \COMMENT{translation vector}
  \FOR{$i \gets 1$ to $k$}
  \STATE $z^{(i)} \gets z + \delta_i \mathbf{v} $ \COMMENT{translation of $z$}
  \STATE $z^{(i)} \gets \mathbb{I}[z^{(i)} > 0.5]$ \COMMENT{optional: binarise vector}
  \ENDFOR
  \STATE \textbf{return} $\{z^{(i)}\}_{i=0}^{k}$
\end{algorithmic}
\end{algorithm}
\end{minipage}

\subsection{Multimodal \ours{} details}\label{app:im2im}

To build the multimodal framework—specifically the image-to-image model composed of \ours{} modules—we train each component independently and evaluate its performance in isolation. This modular approach allows us to assess the effectiveness of each part before assembling the full model, ensuring that all components function reliably. Below, we describe this process for the multimodal \ours{} trained for Edges2Handbags sketch-to-photo task.
\paragraph{\ours{} modules}

\begin{wrapfigure}[13]{r}{0.35\linewidth}
    \centering
    \vspace*{-1.5\baselineskip}
    \includegraphics[width=\linewidth]{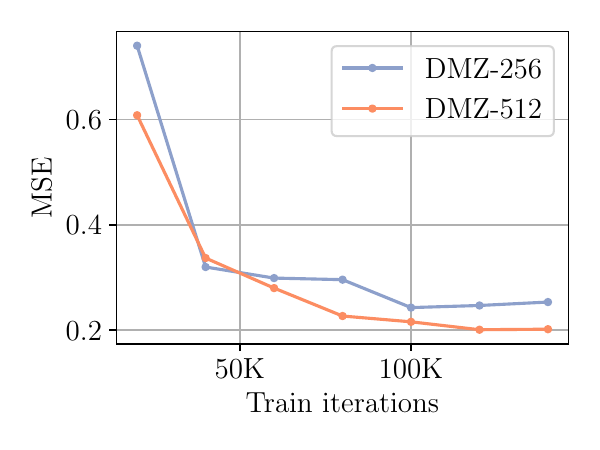}
    \vspace*{-1.5\baselineskip}
    \caption{Reconstruction error of \ours{} models trained on Handbags-64 measured on 10K images, an upper bound for Edges2Handbags task.}
    \label{fig:photo-mse}
\end{wrapfigure}

We use two instances of \ours{}-512: one trained on Edges-64 and the other on Handbags-64. In the sketch-to-photo task, only the model trained on photos is used to generate images, while the model trained on sketches is used to encode their representations. The mean squared error (MSE), defined as $\parallel x_{\text{photo}} - \widehat{x}_{\text{photo}}\parallel$, where $\widehat{x}_{\text{photo}} \sim p_\theta(x_{\text{photo}} | z_{\text{photo}})$ and $z_{\text{photo}} \sim q_\varphi(z | x_{\text{photo}})$, serves as an upper bound of the MSE for the sketch-to-photo generation task. We monitor this metric during training and stop once it no longer improves. Additionally, the latent dimensionality $|z| = 512$ was selected based on that MSE performance.
\Cref{fig:photo-mse} shows the reconstruction error over the course of training.

\paragraph{Mapping $\gamma$}

We train an MLP to learn a mapping $\gamma: Z_{\text{sketch}} \to Z_{\text{photo}}$ using latent codes from the \ours{} models. To determine the optimal architecture, we experiment with different numbers of layers $L$ in the MLP and evaluate MSE $\parallel x_{\text{photo}} - \widehat{x}_{\text{photo}}\parallel$, where $\widehat{x}_{\text{photo}} \sim p_\theta(x_{\text{photo}} \mid \gamma(z_{\text{sketch}}))$ and $z_{\text{sketch}} = q_\varphi(z \mid x_{\text{sketch}})$. The resulting MSEs for $L=1,2,4,6,8$ are $0.26$, $0.24$, $0.23$, $0.22$, and $0.23$, respectively, leading us to select $L=6$ as the optimal depth.
However, note that a simpler mapping would provide greater interpretability for the framework.

\paragraph{Additional capabilities of \ours{} image-to-image framework}

\begin{figure}[htb]
    \centering
    \begin{subfigure}[t]{0.33\linewidth}
        \centering
        \includegraphics[height=50px]{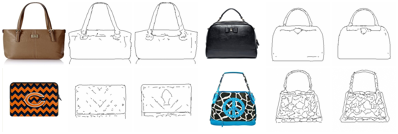}
        \caption{Photo-to-sketch translations}
    \end{subfigure}
    \begin{subfigure}[t]{0.66\linewidth}
        \centering
        \includegraphics[height=50px]{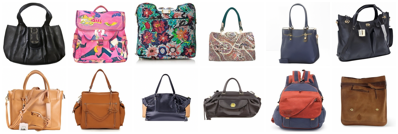}
        \includegraphics[height=50px]{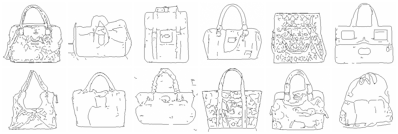}
        \caption{Unconditional generation}
    \end{subfigure}
    \caption{Qualitative results showing additional capabilities of the \ours{} image-to-image framework.}
    \label{fig:im2im}
    \vspace{0.5\baselineskip}
\end{figure}

With our \ours{} framework, we can perform reverse image-to-image mapping---generating sketches from photos---as well as unconditional generation of both photos and sketches. Examples are shown in \Cref{fig:im2im}.
Note that we use PixelSNAIL for unconditional generation, as the latent size $|z|=512$ was chosen to optimise reconstruction loss rather than efficient sampling of $z$.

\paragraph{Evaluation on the test set}

In \cref{tab:im2im} we follow \citet{ zhou2023denoisingdiffusionbridgemodels} and evaluate sketch-to-photo translation on the training set.
The Edges2Handbags test set has only 200 images, too few for stable FID and IS, and the baseline papers do not report test-set results under this protocol, so a direct comparison on the test set is not possible.
As a consequence, this evaluation compares methods under a shared protocol and does not measure generalisation to unseen sketches.
For completeness, we evaluate \ours{} on the 200 test images with the same setting ($T=40$).
We obtain LPIPS $=0.469$ and MSE $=0.446$, compared with $0.359$ and $0.209$ on the training set.
We omit FID and IS because 200 samples are insufficient for reliable estimates.
Both metrics are worse on the test set, as expected for unseen sketches.
As no baseline reports test-set numbers, we cannot place these values relative to prior methods, and include them to make the generalisation gap explicit.

\subsection{Reconstruction Performance of Discrete vs Continuous Representations}\label{app:contin}

\begin{wrapfigure}{r}{0.35\linewidth}
    \centering
    \vspace{-1.5\baselineskip}
    \includegraphics[width=\linewidth]{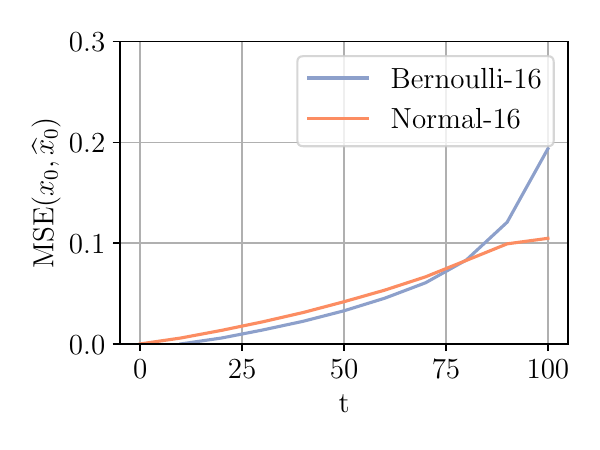}
    \vspace{-2\baselineskip}
    \caption{Reconstruction performance comparison between \ours{} (Bernoulli-16) and continuous latent \( z \) (Normal-16). Mean Squared Error (MSE) is reported on CIFAR-10 for predicting \( x_0 \) from \( x_t \) over \( t \) denoising steps, with \( T=100 \).}
    \label{fig:recon}
\end{wrapfigure}

A continuous latent variable $z$ generally carries more information about the target $x_0$ than a discrete latent of the same size, enabling better reconstruction.
However, this advantage holds primarily when $z$ is the only available information about the target---specifically, at the final timestep $T$. For intermediate noisy inputs $x_t$ with $t < T$, \Cref{fig:recon} shows that the reconstruction error becomes comparable between models with discrete and continuous latents.

The core idea behind DMZ is to learn latent representations that provide the denoiser with additional, useful information to improve denoising efficiency, while keeping the latent space simple enough to allow easy sampling during inference. Importantly, in DMZ, the latent $z$ is not intended to be sufficient for full image reconstruction on its own. Rather, it serves as a complementary guide for the denoiser.
For downstream tasks requiring higher reconstruction fidelity, such as targeted edits or interpolations, one can utilise intermediate states $x_t$. Our main experiments assume $t = T$ to isolate the effect of $z$ alone. \Cref{fig:inter-class} illustrates examples of edits that leverage state $x_t$ to better preserve image identity.

\subsection{Finetuning HuggingFace models}\label{app:finetune}
We have demonstrated that unconditional DDPMs can be effectively finetuned into the \ours{} framework, enabling representation learning, conditional generation, and other capabilities. Here, we finetune the publicly available DDPM model trained on CelebA-HQ ($256\times256$) \citep{karras2018progressive}, available via HuggingFace\footnote{\url{https://huggingface.co/google/ddpm-ema-celebahq-256/}}, into \ours{}.
We train our models for 40K training iterations by finetuning all parameters. Quantitative results are presented in \Cref{tab:hq}.

\begin{table}[htbp]
\centering
\caption{Evaluation of DDPM trained on CelebA-HQ and \ours{} finetuned from it.}
\label{tab:hq}
\begin{tabular}{lccccccc}
\toprule
\multirow{2}{*}{Model} & \multirow{2}{*}{\shortstack{NLL\\(BPD)}}& \multirow{2}{*}{AUROC} & \multicolumn{4}{c}{FID@10K} & IS@10K \\
\cmidrule(lr){4-7} \cmidrule(lr){8-8}
& & & T=10 & T=20 & T=50 & T=100 & T=100 \\
\midrule
DDPM     & 6.25 & --- & 71.43 & 53.55 & 36.86 & 29.81 & 2.77 \\
\ours{}-64  & 3.01 & 74.06$\pm$0.14  & 39.91 &  28.16 & 19.60 &  15.15 & 3.06 \\
\ours{}-256  & 3.00 & 88.53$\pm$0.05 & 49.53 & 42.31 & 33.25 & 27.54 & 2.59 \\
\bottomrule
\end{tabular}
\end{table}

\subsection{\ours{} with Diffusion Transformers}\label{app:dit}

To demonstrate the versatility of the \ours{} framework, we apply it to the attention-based DiT architecture \citep{Peebles2022DiT}. We adapt the model by replacing self-attention layers in every second DiT block with cross-attention layers that attend to a latent $z$. Additionally, we provide results with \ours{} with AdaLN conditioning which is a commonly used mechanism in conditional DiTs. In the AdaLN~\citep{Peebles2022DiT}, $z$ is mapped by an MLP and added to the timestep embedding. The sum determines the per-block scale, shift and gate parameters.

We train DiT-B/4 models on CIFAR-10 for 200K iterations without specific hyperparameter tuning\footnote{\url{https://github.com/facebookresearch/DiT}}. As shown in Table~\ref{tab:dit_results}, \ours{} significantly improves generative quality at lower inference ($T=10,20$) steps compared to the baseline DiT, while maintaining an identical NLL. 
Both conditioning strategies substantially outperform the unconditional baseline at small step budgets. Cross-attention has a slight edge at $T=20,50$, while AdaLN is slightly better at $T=10,100$. Crucially, there does not appear to be a clear difference between the two approaches. This suggests that the benefit of $z$ generalises across architectures and conditioning mechanisms.
Furthermore, the learned representations achieve higher classification accuracy than the U-Net-based results reported in Table~\ref{tab:acc}.

\begin{table}[h]
    \centering
    \vspace{-0.5\baselineskip}
    \caption{Comparison of FID scores on CIFAR-10 between DiT and DiT adapted with \ours{}.}
    \label{tab:dit_results}
    \begin{tabular}{lcccccccc}
        \toprule
         \multirow{2}{*}{Model} & \multirow{2}{*}{Conditioning} & \multirow{2}{*}{\shortstack{NLL\\(BPD)}} & \multirow{2}{*}{Acc} & \multicolumn{4}{c}{FID@50K} & IS@50K \\
         \cmidrule(lr){5-8} \cmidrule(lr){9-9}
        & & & & $T=10$ & $T=20$ & $T=50$ & $T=100$ & $T=100$\\
        \midrule
        DiT-B/4 & -- & 3.30 & -- & 32.14 & 15.56 & 10.79 & 10.15 & 8.52 \\
        \ours{}-16  & Cross-attention & 3.30 & 51.52 & 17.76 & 12.28 & 10.81 & 10.91 & 8.32 \\
        \ours{}-16 & AdaLN & 3.30 & 51.08 & 17.63 & 12.36 & 11.86 & 10.89 & 8.31 \\
        \bottomrule
    \end{tabular}
\end{table}

\subsection{\ours{} with ImageNet}
Additionally, we train \ours{} and baseline DDPM models on ImageNet $32\times 32$ \citep{5206848,imagenet15russakovsky} to evaluate performance on a larger-scale dataset (1.28M images, 1K classes). We use the same training setup as CIFAR-10 due to the identical resolution, without further hyperparameter tuning. Results are reported in \Cref{tab:imagenet}.

\begin{table}[h] 
  \centering
  \vspace{-0.5\baselineskip}
  \caption{Evaluation of \ours{} trained on ImageNet ($32\times 32$).} 
  \label{tab:imagenet}
  \begin{tabular}{@{}lcccccccc}
    \toprule
   \multirow{2}{*}{Model} & \multirow{2}{*}{\shortstack{train\\iter.}} & \multirow{2}{*}{\shortstack{NLL\\(BPD)}} & \multirow{2}{*}{Acc} & \multicolumn{4}{c}{\hspace{-8pt}FID@50K} & IS@50K \\
         \cmidrule(lr){5-8} \cmidrule(lr){9-9}
        & & & & $T=10$ & $T=20$ & $T=50$ & $T=100$ & $T=100$\\
    \midrule
    DDPM &  850K & 4.15 & -- & 37.04 & 13.14 & 6.56 & 5.75 & 12.14 \\
    \ours{}-16 & 450K & 4.13 & 9.34 & 19.30 & 8.54 & 6.22 & 6.17 & 12.13 \\
    \ours{}-32 & 450K & 4.13 & 11.25 & 17.67 & 10.22 & 8.16 & 7.82 & 11.25 \\
    \ours{}-64 & 450K & 4.12 & 8.48 & 16.63 & 12.67 & 11.73 & 12.02 & 10.32 \\
    \bottomrule
  \end{tabular}
  \vspace*{0\baselineskip}
\end{table}

\subsection{Additional scores and samples}\label{app:fid10}

\begin{figure}[htb!]
    \centering
    \begin{subfigure}[b]{0.3\linewidth}
        \centering
        \includegraphics[trim=5mm 6mm 5mm 5mm, clip, width=\linewidth]{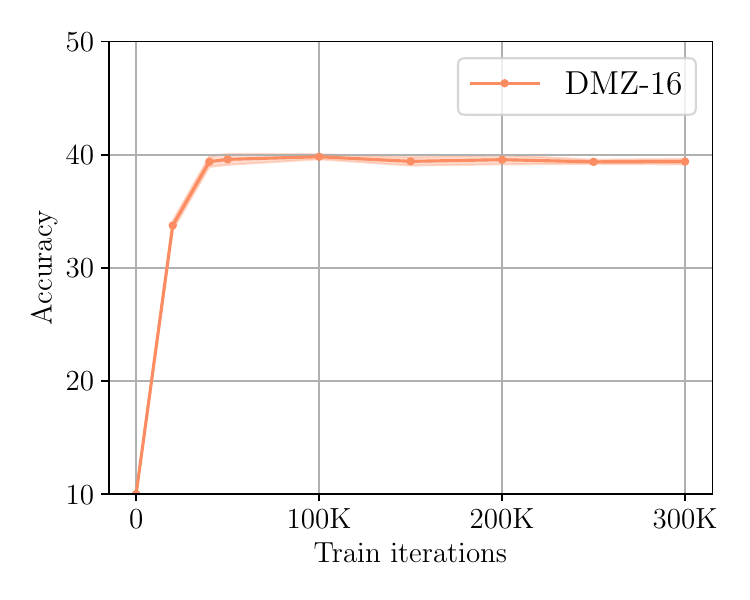}
        \caption{CIFAR-10}
    \end{subfigure}
    \hspace{4mm}
    \begin{subfigure}[b]{0.3\linewidth}
        \centering
        \includegraphics[trim=5mm 6mm 5mm 5mm, clip, width=\linewidth]{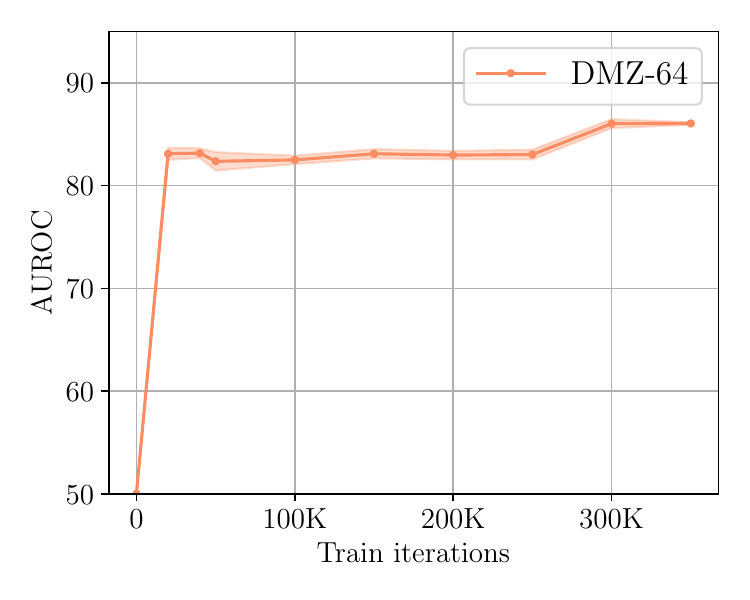}
        \caption{CelebA-64}
    \end{subfigure}
    \caption{Evolution of representation quality during training. High-quality representations emerge early and remain stable throughout.}
    \label{fig:acc}
\end{figure}

\begin{figure}[htbp]
    \centering
    \begin{subfigure}[b]{\textwidth}
        \centering
        \begin{subfigure}[b]{0.32\textwidth}
            \includegraphics[trim=70 2 0 1, clip, width=\linewidth]{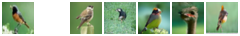}
            \includegraphics[trim=70 2 0 1, clip, width=\linewidth]{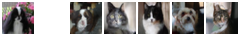}
            \includegraphics[trim=70 2 0 1, clip, width=\linewidth]{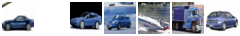}
            \includegraphics[trim=70 2 0 1, clip, width=\linewidth]{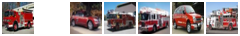}
            \caption*{$|z|=16$}
        \end{subfigure}
        \hfill
        \begin{subfigure}[b]{0.32\textwidth}
            \includegraphics[trim=70 2 0 1, clip, width=\linewidth]{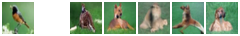}
            \includegraphics[trim=70 2 0 1, clip, width=\linewidth]{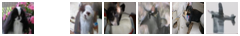}
            \includegraphics[trim=70 2 0 1, clip, width=\linewidth]{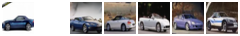}
            \includegraphics[trim=70 2 0 1, clip, width=\linewidth]{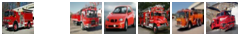}
            \caption*{$|z|=32$}
        \end{subfigure}
        \hfill
        \begin{subfigure}[b]{0.32\textwidth}
            \includegraphics[trim=70 2 0 1, clip, width=\linewidth]{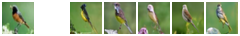}
            \includegraphics[trim=70 2 0 1, clip, width=\linewidth]{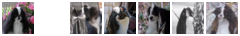}
            \includegraphics[trim=70 2 0 1, clip, width=\linewidth]{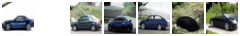}
            \includegraphics[trim=70 2 0 1, clip, width=\linewidth]{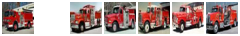}
            \caption*{$|z|=64$}
        \end{subfigure}
        \caption{CIFAR-10}
    \end{subfigure}

    \vspace{1em} 

    \begin{subfigure}[b]{\textwidth}
        \centering
        \begin{subfigure}[b]{0.32\textwidth}
            \includegraphics[trim=134 2 0 1, clip, width=\linewidth]{imgs/examples/celeba64_keep_z_sample_x_2_64.png}
            \includegraphics[trim=134 2 0 1, clip, width=\linewidth]{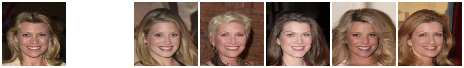}
            \includegraphics[trim=134 2 0 1, clip, width=\linewidth]{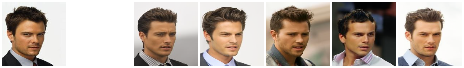}
            \includegraphics[trim=134 2 0 1, clip, width=\linewidth]{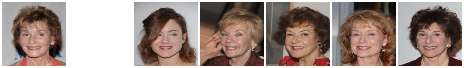}
            \caption*{$|z|=64$}
        \end{subfigure}
        \hfill
        \begin{subfigure}[b]{0.32\textwidth}
            \includegraphics[trim=134 2 0 1, clip, width=\linewidth]{imgs/examples/celeba64_keep_z_sample_x_2_128.png}
            \includegraphics[trim=134 2 0 1, clip, width=\linewidth]{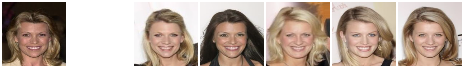}
            \includegraphics[trim=134 2 0 1, clip, width=\linewidth]{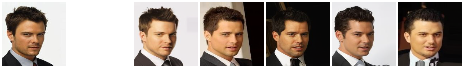}
            \includegraphics[trim=134 2 0 1, clip, width=\linewidth]{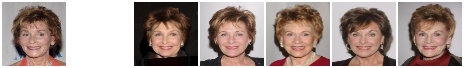}
            \caption*{$|z|=128$}
        \end{subfigure}
        \hfill
        \begin{subfigure}[b]{0.32\textwidth}
            \includegraphics[trim=134 2 0 1, clip, width=\linewidth]{imgs/examples/celeba64_keep_z_sample_x_2_256.png}
            \includegraphics[trim=134 2 0 1, clip, width=\linewidth]{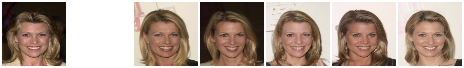}
            \includegraphics[trim=134 2 0 1, clip, width=\linewidth]{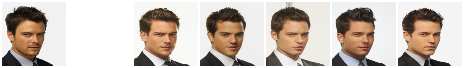}
            \includegraphics[trim=134 2 0 1, clip, width=\linewidth]{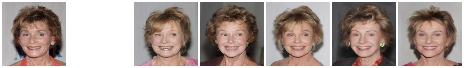}
            \caption*{$|z|=256$}
        \end{subfigure}
        \caption{CelebA-64}
    \end{subfigure}

    \caption{Comparison of representations learned by \ours{} on CIFAR-10 and CelebA-64 for varying latent sizes $|z|$. Images are generated from $z \sim q_\phi(z \mid x_0),\ x_0 \sim \mathcal{D}$ and five different $x_T \sim \mathcal{N}(0, \mathbf{I})$.}
    \label{fig:samples-rep}
\end{figure}

\begin{figure}[htbp]
    \centering
    \begin{subfigure}[b]{0.48\textwidth}
         \includegraphics[width=\linewidth]{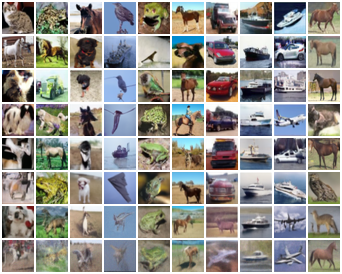}
        \caption{\ours{}-16 for CIFAR-10}
    \end{subfigure}
    \hfill
    \begin{subfigure}[b]{0.48\textwidth}
        \includegraphics[width=\linewidth]{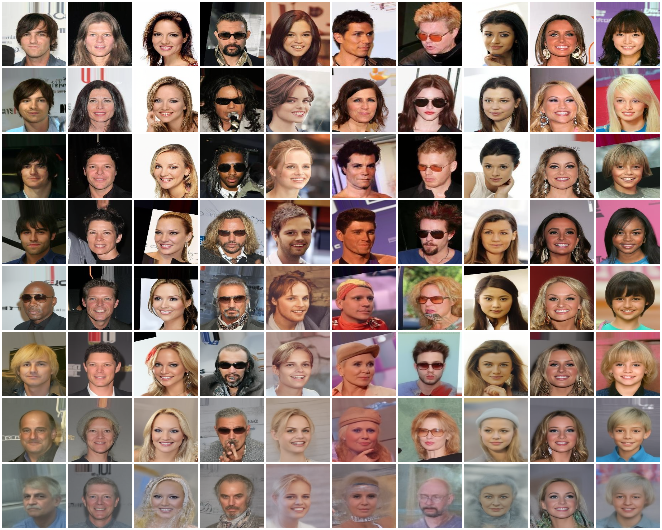}
        \caption{\ours{}-64 for CelebA-64}
    \end{subfigure}

    \caption{Images generated with varying numbers of denoising steps $T$. Each column shows samples generated from a fixed latent code $z \sim \text{Bernoulli}$. Rows correspond to $T = 1000, 500, 200, 100, 50, 20, 10, 5$ steps, from top to bottom.}
    \label{fig:samples}
\end{figure}

\begin{figure}[htbp]
    \centering
    \begin{subfigure}[b]{0.48\textwidth}
         \includegraphics[width=\linewidth]{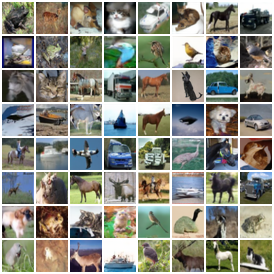}
        \caption{\ours{}-16 for CIFAR-10}
    \end{subfigure}
    \hfill
    \begin{subfigure}[b]{0.48\textwidth}
        \includegraphics[width=\linewidth]{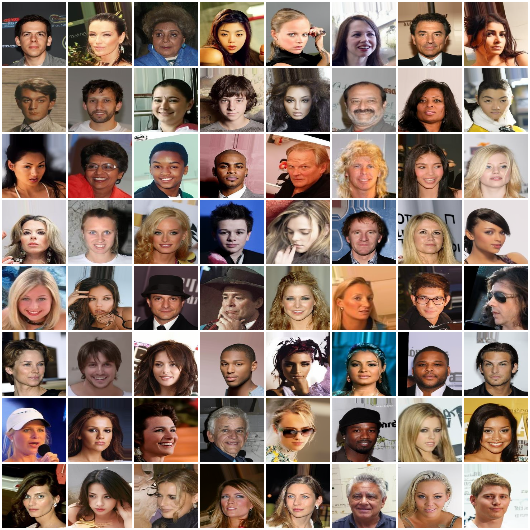}
        \caption{\ours{}-64 for CelebA-64}
    \end{subfigure}
    \hfill
    \begin{subfigure}[b]{\textwidth}
        \vspace*{2mm}
        \includegraphics[width=\linewidth]{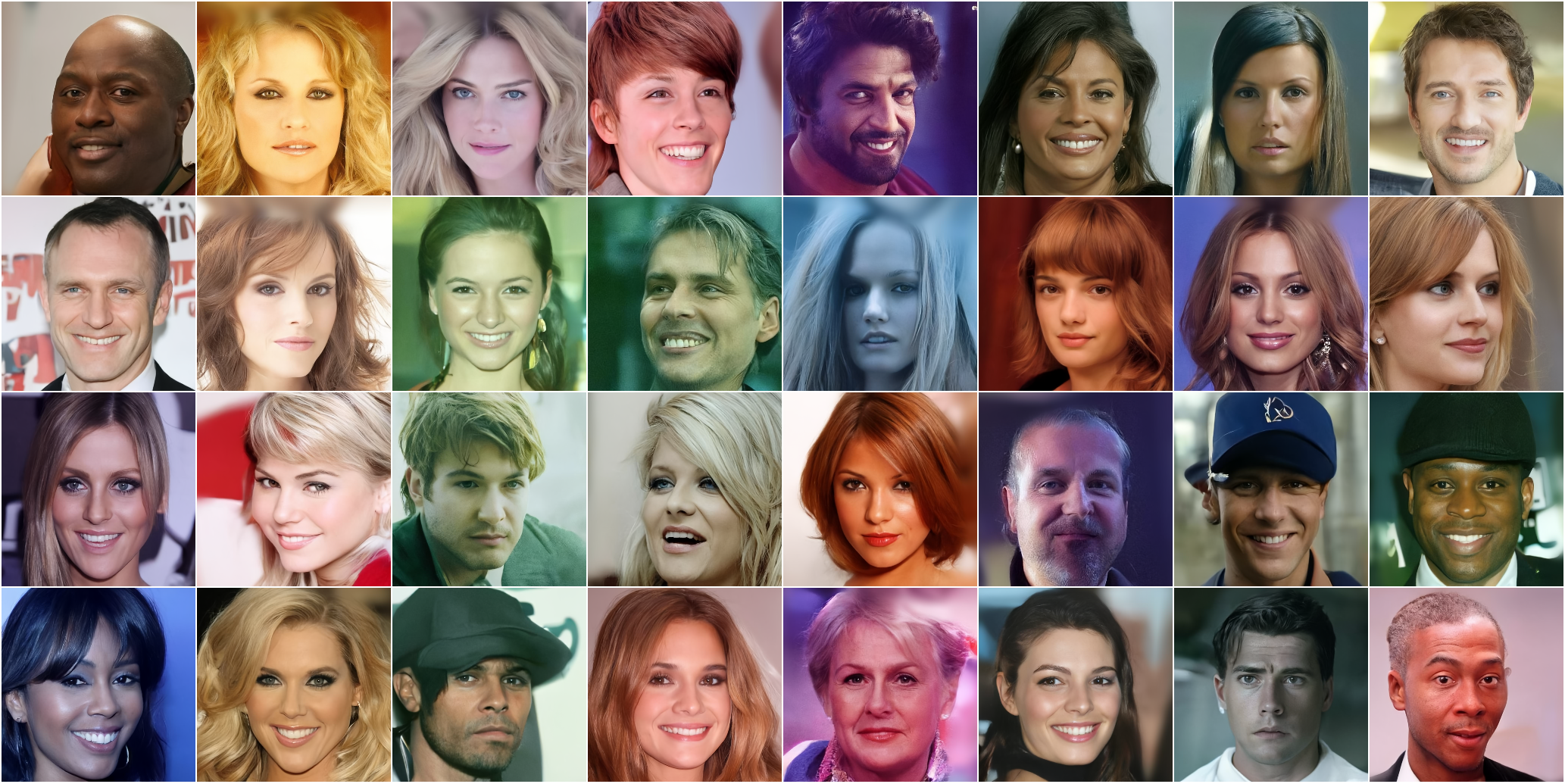}
        \caption{\ours{}-64 for CelebA-HQ}
    \end{subfigure}
    \caption{Generated sample images produced using $T=100$ diffusion steps.}
    \label{fig:gen}
\end{figure}

Additional scores are reported in \Cref{tab:fid10} (FID@10K) \Cref{tab:is} (Inception Scores) \Cref{tab:more-metrics} (reconstruction metrics), \Cref{fig:acc} (accuracy over training iterations), and additional examples are provided in \Cref{fig:samples,fig:samples-rep,fig:gen}.

\begin{table}[htbp]
    \centering
    \vspace{-0.5\baselineskip}
    \caption{FID scores for \ours{} models evaluated with 10K samples. Reported values are the average $\pm$ standard deviation over 5 models.}
    \label{tab:fid10}
    \begin{tabular}{ccccccccc}
    \toprule
     Model & Dataset & T=10 & T=20 & T=50 & T=100 \\
     \midrule
     \ours{}-16 & CIFAR-10 & 11.53$\pm$0.90 & 6.76$\pm$0.69 & 4.96$\pm$0.36 & 4.67$\pm$0.24 \\
     \ours{}-64 & CelebA-64 & 15.70$\pm$0.50 & 8.91$\pm$0.38 & 5.18$\pm$0.27 & 3.84$\pm$0.48 \\
     \bottomrule
    \end{tabular}
\end{table}

\begin{table}[ht]
  \centering
  \caption{Inception Scores (IS) measured for 50K generated samples. Note that DiffuseVAE is the only baseline that evaluates IS.}
   \label{tab:is}
    \small
    \begin{tabular}{@{}l@{}rrr}
        \toprule
         Model & $T$ & CIFAR-10 & CelebA-64 \\
        \midrule
        \multirow{4}{*}{\ours} & 10 & 9.10$\pm$0.08 & 3.33$\pm$0.12\\
        & 20 & 9.39$\pm$0.08 & 3.36$\pm$0.10 \\
        & 50 &  9.59$\pm$0.14 &  3.32$\pm$0.08 \\
        & 100 & 9.64$\pm$0.07 & 3.33$\pm$0.10 \\
        \midrule
        \multirow{4}{*}{\shortstack[l]{DDPM\\ \citep{ning2023inputperturbationreducesexposure} \\(reproduced)}} & 10 & 8.36 & 2.82 \\ 
        & 20 & 9.19 & 3.94 \\
        & 50 & 9.38 & 3.01 \\
        & 100 & 9.42 & 3.10\\
        \midrule
        DiffuseVAE \citep{pandey2022diffusevaeefficientcontrollablehighfidelity} \hspace{1pt} & 1000  & 9.60 & -- \\
        \bottomrule
    \end{tabular}
\end{table}

\begin{table}[ht]
    \centering
    \caption{Reconstruction metrics achieved by \ours{}. Comparison against the test set.}
    \label{tab:more-metrics}
    \small
    \begin{tabular}{lrrrrrrr}
    \toprule
    Dataset & $|z|$ & SSIM & LPIPS & MSE & PSNR & IS & FID \\
    \midrule
    \multirow{4}{*}{CIFAR-10} & 16 & 0.090 & 0.624 & 0.378 & 10.277 & 9.705 & 4.618\\
    & 32 & 0.119 & 0.596 & 0.347 & 10.647 & 9.639 & 4.821 \\
    & 64 & 0.225 & 0.556 & 0.248 & 12.150 & 9.529 & 4.798 \\
    & 128 & 0.374 & 0.510 & 0.139 & 14.712 & 9.395 & 4.763 \\
    \midrule
    \multirow{4}{*}{CelebA-64} & 32 & 0.106 & 0.541 & 0.534 & 8.742 & 3.137 & 3.864 \\
    & 64 & 0.145 & 0.493 & 0.483 & 9.183 & 3.223 & 3.170 \\
    & 128 & 0.224 & 0.447 & 0.405 & 9.953 & 3.294 & 3.593 \\
    & 256 & 0.324 & 0.403 & 0.298 & 11.298 & 3.264 & 3.092\\
    \bottomrule
    \end{tabular}
\end{table}

\subsection{Reproducibility Details}

We adopt the hyperparameter settings from \citet{ning2023inputperturbationreducesexposure}, which are based on the configurations by \citet{dhariwal2021diffusionmodelsbeatgans}. The specific values are listed in \cref{tab:hparams}. All models are trained using the AdamW optimiser \citep{loshchilov2019decoupledweightdecayregularization} with 16-bit mixed precision training with loss scaling \citep{micikevicius2018mixedprecisiontraining,dhariwal2021diffusionmodelsbeatgans}, while keeping the model weights, EMA, and optimiser states in 32-bit precision. An EMA decay rate of 0.9999 is used in all experiments, following \citet{ning2023inputperturbationreducesexposure} setup.

The encoder used to extract codes $z$ from input images consists of repeated blocks of a convolutional layer, batch normalisation, and LeakyReLU activation, followed by a final projection layer. The encoder follows a simple resolution-dependent architecture rule. We use 4 blocks for $32 \times 32$ images, 5 blocks for $64 \times 64$, and 7 blocks for $256 \times 256$.


We use PyTorch~\citep{paszke2019pytorchimperativestylehighperformance}, and train all models with Python 3.10 and PyTorch version 2.6.
For CIFAR-10, we use a single NVIDIA A40 GPU and train for 1-2 days. For CelebA-64, we use two A40 GPUs and train for about 10 days.
Models trained on Edges2Handbags—handled separately as Edges and Handbags—are also trained using two A40 GPUs, with a training time of around 3 days.
Finetuning of CelebA-HQ DDPM~\citep{ho2020denoisingdiffusionprobabilisticmodels} takes slightly less than 2 days on 4 A40 GPUs.
For T=100, sampling a batch of 64 images using a single A40 GPU takes 12s, 264.5s, 79.3s, 79.3s, and 222.5s, for CIFAR-10, CelebA, Edges, Handbags, and CelebA-HQ, respectively.

Our implementation and instructions for reproducing the experiments are available at \url{https://github.com/exlab-research/dmz}.

\begin{table*}[htb!]
\caption{Hyperparameter values based on \citet{ning2023inputperturbationreducesexposure} for all datasets except CelebA-HQ, where we follow the configuration from \citet{ho2020denoisingdiffusionprobabilisticmodels} and perform finetuning only.}
\label{tab:hparams}
\begin{center}
\begin{tabular}{@{}llllll@{}}
\toprule
\centering
 &  CIFAR-10 & CelebA & Edges & Handbags & CelebA-HQ$^*$  \\
 & $32\times32$ & $64\times64$ & $64\times64$ & $64\times64$ & $256\times256$ \\
 \midrule
Size of $z$ & 16/32/64 & 64/128/256 & 512 & 512 & 64/256 \\
Training diffusion steps $T_{\text{train}}$ & 1,000 & 1,000 & 1,000 & 1,000 & 1,000 \\
UNet size & 68M & 409M & 333M & 333M & 142M \\
Encoder size & 0.5M & 1.8M & 3.7M & 3.7M & 29M \\
Encoder channels & 32, 64, 128, 256 & 32, \dots, 512  & 32, \dots, 512 & 32, \dots, 512 & 32, \dots, 2048 \\

Channels & 128  & 192  & 192 & 192  & 128 \\
Residual blocks & 3 & 3 & 3 & 3 & 2 \\
Channels multiple & 1, 2, 2, 2 & 1, 2, 3, 4 & 1, 2, 3, 4 & 1, 2, 3, 4 & 1, 1, 2, 2, 4, 4\\
Heads channels & 32 & 64 & 64 & 64 & 512 \\
Attention resolution & 16, 8 & 32, 16, 8 & 32, 16, 8 & 32, 16, 8 & 16 \\
Cross attention resolution & 16, 8 & 32, 16, 8 & 16 & 16 & 16 \\
Mid-block cross attention & True & True & True & True & True \\
BigGAN up/downsample & True & True & True & True & True \\
Dropout & 0.3 & 0.1 & 0.1 & 0.1 & 0.1 \\
Batch size & 128 & 256 & 256 & 256 & 256 \\
Training iterations & 250K & 300K & 120K & 120K & 40K \\
Training images & 50K & 163K & 139K & 139K & 24K \\
Learning rate & 1e-4 & 1e-4 & 1e-4 & 1e-4 & 1e-4 \\
Learned sigma & True & True & True & True & False \\
Noise schedule & cosine & cosine & cosine & cosine & linear \\
Input perturbation & 0.15 & 0.1 & 0.1 & 0.1 & 0.1 \\
\bottomrule
\end{tabular}
\end{center}
\end{table*}

\end{document}

%% file: math_commands.tex
\usepackage{amsmath,amsfonts,bm}

\def\eqref#1{equation~\ref{#1}}

\def\1{\bm{1}}

\DeclareMathAlphabet{\mathsfit}{\encodingdefault}{\sfdefault}{m}{sl}
\SetMathAlphabet{\mathsfit}{bold}{\encodingdefault}{\sfdefault}{bx}{n}



%% file: imgs/diagrams/set2colors.tex
\definecolor{set2green}{rgb}{0.627, 0.816, 0.180}
\definecolor{set2pink}{rgb}{0.550, 0.212, 0.494}
\definecolor{set2blue}{rgb}{0.329, 0.649, 0.827}
\definecolor{set2orange}{rgb}{0.780, 0.358, 0.141}
\definecolor{set2lightgreen}{rgb}{0.698, 0.626, 0.173}
\definecolor{set2teal}{rgb}{0.176, 0.592, 0.466} 
\definecolor{set2brown}{rgb}{0.835, 0.333, 0.238} 
\definecolor{set2grey}{rgb}{0.557, 0.229, 0.505}

%% file: imgs/diagrams/diff-hvi.tex
\begin{tikzpicture}[
    every node/.style={inner sep=0pt},
    var/.style={draw,circle,thick,minimum size=24pt},
    destruction/.style={-latex, ultra thick},
    generation/.style={-latex, ultra thick, dashed},
    y=48pt,x=72pt
    ]

\pgfdeclarelayer{background}
\pgfsetlayers{background,main}

\node[var](x0) at (0.5,0) {$x_0$};
\node[var](xprev) at (2,0) {$x_{t-1}$};
\node[var](xt) at (3,0) {$x_t$};
\node[var,fill=set2blue!20](xT) at (4.5,0) {$x_T$};
\node[var,draw=none](x1) at (1.25,0) {$\cdots$};
\node[var,draw=none](xrest) at (3.75,0) {$\cdots$};

\draw[destruction] (x0) -- (x1);
\draw[destruction] (x1) -- (xprev);
\draw[destruction] (xprev) to[bend left=30] node[midway,above,yshift=2pt] {$q(x_t\mid x_{t-1})$} (xt);
\draw[destruction] (xt) -- (xrest);
\draw[destruction] (xrest) -- (xT);

\node[var,dashed](xhat) at (2,-1) {$\hat{x}_0$};
\draw[generation] (xhat) -- (xprev) node[pos=0.8](toxprev) {};
\draw[generation,draw=set2brown] (xt) -- (xhat) node[pos=0.2](fromxt) {};

\begin{pgfonlayer}{background}
    \node[fit=(xhat)(xprev.west)(xt.east), inner ysep=-7pt, inner xsep=5pt, yshift=-15pt,rounded corners=5pt,fill=set2blue!20, thick, draw=black,align=right]
    (block) {};
\end{pgfonlayer}

\node[anchor=south east,inner sep=5pt] at (block.south east) {$p_\theta(x_{t-1}\mid x_t)$};

\end{tikzpicture}

%% file: imgs/diagrams/diff-learnable.tex
\begin{tikzpicture}[
    every node/.style={inner sep=0pt},
    var/.style={draw,circle,thick,minimum size=24pt},
    destruction/.style={-latex, ultra thick},
    generation/.style={-latex, ultra thick, dashed},
    y=48pt,x=72pt
    ]

\pgfdeclarelayer{background}
\pgfsetlayers{background,main}

\node[var](x0) at (0.5,0) {$x_0$};
\node[var](xprev) at (2,0) {$x_{t-1}$};
\node[var](xt) at (3,0) {$x_t$};
\node[var,fill=set2blue!20](xT) at (4.5,0) {$x_T$};
\node[var,draw=none](x1) at (1.25,0) {$\cdots$};
\node[var,draw=none](xrest) at (3.75,0) {$\cdots$};

\draw[destruction] (x0) -- (x1);
\draw[destruction] (x1) -- (xprev);
\draw[destruction,draw=set2brown] (xprev) to[bend left=30] (xt);
\draw[destruction,draw=set2brown] (x0) to[bend left=30] node[midway,above,yshift=2pt] {$q_\phi(x_t\mid x_0,x_{t-1})$} (xt);
\draw[destruction] (xt) -- (xrest);
\draw[destruction] (xrest) -- (xT);

\node[var,dashed](xhat) at (2,-1) {$\hat{x}_0$};
\draw[generation] (xhat) -- (xprev) node[pos=0.8](toxprev) {};
\draw[generation,draw=set2brown] (xt) -- (xhat) node[pos=0.2](fromxt) {};

\begin{pgfonlayer}{background}
    \node[fit=(xhat)(xprev.west)(xt.east), inner ysep=-7pt, inner xsep=5pt, yshift=-15pt,rounded corners=5pt,fill=set2blue!20, thick, draw=black,align=right]
    (block) {};
\end{pgfonlayer}

\node[anchor=south east,inner sep=5pt] at (block.south east) {$p_\theta(x_{t-1}\mid x_t)$};

\end{tikzpicture}

%% file: imgs/diagrams/diff-dmz.tex
\begin{tikzpicture}[
    every node/.style={inner sep=0pt},
    var/.style={draw,circle,thick,minimum size=24pt},
    destruction/.style={-latex, ultra thick},
    generation/.style={-latex, ultra thick, dashed},
    y=48pt,x=72pt
    ]

\pgfdeclarelayer{background}
\pgfsetlayers{background,main}

\node[var](x0) at (0.5,0) {$x_0$};
\node[var](xprev) at (2,0) {$x_{t-1}$};
\node[var](xt) at (3,0) {$x_t$};
\node[var,fill=set2blue!20](xT) at (4.5,0) {$x_T$};
\node[var,draw=none](x1) at (1.25,0) {$\cdots$};
\node[var,draw=none](xrest) at (3.75,0) {$\cdots$};

\draw[destruction] (x0) -- (x1);
\draw[destruction] (x1) -- (xprev);
\draw[destruction] (xprev) to[bend left=30] node[midway,above,yshift=2pt] {$q(x_t\mid x_{t-1})$} (xt);
\draw[destruction] (xt) -- (xrest);
\draw[destruction] (xrest) -- (xT);

\node[var,dashed](xhat) at (2,-1) {$\hat{x}_0$};
\draw[generation] (xhat) -- (xprev) node[pos=0.8](toxprev) {};
\draw[generation,-,draw=set2brown] (xt) to[bend left=30] ([xshift=24pt]xhat.east) node[pos=0.2](fromxt) {};

\begin{pgfonlayer}{background}
    \node[fit=(xhat)(xprev.west)(xt.east), inner ysep=-7pt, inner xsep=5pt, yshift=-15pt,rounded corners=5pt,fill=set2blue!20, thick, draw=black,align=right]
    (block) {};
\end{pgfonlayer}

\node[anchor=south east,inner sep=5pt] at (block.south east) {$p_\theta(x_{t-1}\mid x_t,z)$};

\node[var,fill=set2blue!20](z) at (3.75,-1) {$z$};
\draw[generation,draw=set2brown] (z) -- (xhat);
\draw[destruction,draw=set2brown] (x0) to[bend right=50] node[pos=0.65,below,yshift=-2pt] {$q_\phi(z\mid x_0)$} (z);



\end{tikzpicture}

%% file: imgs/diagrams/uncond.tex
\begin{tikzpicture}[
    every node/.style={inner sep=5pt},
    block/.style={draw, minimum width=150px, minimum height=30pt, thick, rounded corners=10pt},
    flow/.style={-latex, ultra thick},
    y=20pt,x=100pt
    ]

\pgfdeclarelayer{background}
\pgfsetlayers{background,main}

\node[block, fill=set2blue!20](uds) at (0,8) {\Large Up / Downsample};
\node[block, fill=set2brown!20](att) at (0,5.5) {\Large Attention\vphantom{g}};
\node[block, fill=set2blue!20](resnet) at (0,2) {\Large ResNet Block\vphantom{g}};

\draw[flow] (att) -- (uds);

\draw[flow] (resnet.north) -- ++(0,0.75) -- ++(0,0) -- (att.south) node[midway,right](k) {\large \textbf{K}\vphantom{Q}};
\draw[flow] (resnet.north) -- ++(0,0.75) -- ++(-0.3,0) -- ([xshift=-30]att.south) node[midway,left] {\large \textbf{Q}};
\draw[flow] (resnet.north) -- ++(0,0.75) -- ++(0.3,0) -- ([xshift=30]att.south) node[midway,right] {\large \textbf{V}\vphantom{Q}};

\node[anchor=north](prev) at (0,0) {\Large prev.\ layer};
\node[](next) at (0,10.5) {};
\draw[flow] (prev) -- (resnet);
\draw[flow] (uds) -- (next);

\node[anchor=north](t) at (0.5,0) {\Large $t$\vphantom{gl}};
\draw[flow] (t) -- ([xshift=50]resnet.south);

\begin{pgfonlayer}{background}
    \node[fit=(uds)(att)(resnet), inner sep=15pt, rounded corners=10pt, ultra thick,draw=black!50] (block) {};
    \node[fit=(att)(k), inner sep=7.5pt, rounded corners=10pt, ultra thick,fill=yellow!20,dotted,draw=black!50] (attblock) {};
\end{pgfonlayer}

\end{tikzpicture}

%% file: imgs/diagrams/condwitht.tex
\begin{tikzpicture}[
    every node/.style={inner sep=5pt},
    block/.style={draw, minimum width=150px, minimum height=30pt, thick, rounded corners=10pt},
    flow/.style={-latex, ultra thick},
    y=20pt,x=100pt
    ]

\pgfdeclarelayer{background}
\pgfsetlayers{background,main}

\node[block, fill=set2blue!20](uds) at (0,8) {\Large Up / Downsample};
\node[block, fill=set2brown!20](att) at (0,5.5) {\Large Attention\vphantom{g}};
\node[block, fill=set2blue!20](resnet) at (0,2) {\Large ResNet Block\vphantom{g}};

\draw[flow] (att) -- (uds);

\draw[flow] (resnet.north) -- ++(0,0.75) -- ++(0,0) -- (att.south) node[midway,right](k) {\large \textbf{K}\vphantom{Q}};
\draw[flow] (resnet.north) -- ++(0,0.75) -- ++(-0.3,0) -- ([xshift=-30]att.south) node[midway,left] {\large \textbf{Q}};
\draw[flow] (resnet.north) -- ++(0,0.75) -- ++(0.3,0) -- ([xshift=30]att.south) node[midway,right] {\large \textbf{V}\vphantom{Q}};

\node[anchor=north](prev) at (0,0) {\Large prev.\ layer};
\node[](next) at (0,10.5) {};
\draw[flow] (prev) -- (resnet);
\draw[flow] (uds) -- (next);

\node[anchor=north](t) at (0.5,0) {\Large $t$\vphantom{gl}};
\node[anchor=north, set2brown](z) at (0.65,0) {\Large $\textbf{\textit{z}}$\vphantom{gl}};
\draw[flow] (t) -- ([xshift=50]resnet.south);
\draw[flow, set2brown] (z) -- ([xshift=65]resnet.south);

\begin{pgfonlayer}{background}
    \node[fit=(uds)(att)(resnet), inner sep=15pt, rounded corners=10pt, ultra thick,draw=black!50] (block) {};
    \node[fit=(att)(k), inner sep=7.5pt, rounded corners=10pt, ultra thick,fill=yellow!20,dotted,draw=black!50] (attblock) {};
\end{pgfonlayer}

\end{tikzpicture}

%% file: imgs/diagrams/condatt.tex
\begin{tikzpicture}[
    every node/.style={inner sep=5pt},
    block/.style={draw, minimum width=150px, minimum height=30pt, thick, rounded corners=10pt},
    flow/.style={-latex, ultra thick},
    y=20pt,x=100pt
    ]

\pgfdeclarelayer{background}
\pgfsetlayers{background,main}

\node[block, fill=set2blue!20](uds) at (0,8) {\Large Up / Downsample};
\node[block, fill=set2brown!20](att) at (0,5.5) {\Large Attention\vphantom{g}};
\node[block, fill=set2blue!20](resnet) at (0,2) {\Large ResNet Block\vphantom{g}};

\draw[flow] (att) -- (uds);

\node[set2brown](z) at (0.7,3.5) {\Large $\textbf{\textit{z}}$};

\draw[flow,set2brown] (z) -- ++(-0.7,0) -- (att.south) node[midway,right](k) {\large \textbf{K}\vphantom{Q}};
\draw[flow] ([xshift=-30]resnet.north) -- ++(0,0.75) -- ([xshift=-30]att.south) node[midway,left] {\large \textbf{Q}};
\draw[flow,set2brown] (z) -- ++(-0.4,0) -- ([xshift=30]att.south) node[midway,right] {\large \textbf{V}\vphantom{Q}};

\node[anchor=north](prev) at (0,0) {\Large prev.\ layer};
\node[](next) at (0,10.5) {};
\draw[flow] (prev) -- (resnet);
\draw[flow] (uds) -- (next);

\node[anchor=north](t) at (0.5,0) {\Large $t$\vphantom{gl}};
\draw[flow] (t) -- ([xshift=50]resnet.south);

\begin{pgfonlayer}{background}
    \node[fit=(uds)(att)(resnet), inner sep=15pt, rounded corners=10pt, ultra thick,draw=black!50] (block) {};
    \node[fit=(att)(k), inner sep=7.5pt, rounded corners=10pt, ultra thick,fill=yellow!20,dotted,draw=black!50] (attblock) {};
\end{pgfonlayer}

\end{tikzpicture}